\documentclass[conference]{IEEEtran}
\IEEEoverridecommandlockouts
\usepackage{cite}
\usepackage{amsmath,amssymb,amsfonts}
\usepackage{algorithmic}
\usepackage[pdftex]{graphicx}
\usepackage{textcomp}
\usepackage{xcolor}
\def\BibTeX{{\rm B\kern-.05em{\sc i\kern-.025em b}\kern-.08em
    T\kern-.1667em\lower.7ex\hbox{E}\kern-.125emX}}
    
\usepackage{booktabs}       
\usepackage{caption}
\usepackage[labelformat=parens]{subcaption}

\begin{document}

\title{Kernel Normalized Convolutional Networks for Privacy-Preserving Machine Learning
\thanks{
\textit{To appear in the IEEE Conference on Secure and Trustworthy Machine Learning (SaTML), February 2023}.}
}


\author{\IEEEauthorblockN{Reza Nasirigerdeh}
\IEEEauthorblockA{\textit{Technical University of Munich} \\
\textit{Klinikum rechts der Isar} \\ 
Munich, Germany \\
}
\and
\IEEEauthorblockN{Javad Torkzadehmahani}
\IEEEauthorblockA{\textit{Azad University of Kerman}\\
Kerman, Iran \\
}
\and
\IEEEauthorblockN{Daniel Rueckert}
\IEEEauthorblockA{\textit{Technical University of Munich} \\
\textit{Klinikum rechts der Isar} \\ 
Munich, Germany \\
\textit{Imperial College London}\\
London, United Kingdom \\
}
\and
\IEEEauthorblockN{\ \ \ \ \ \ \ \ \ \ \ \ \ \ \ \ \ \ \ \ \  \ \ \ \ \  \ \ \ \ \  \ \ \ \ \ \ \ }
\IEEEauthorblockA{ \ \ \ \ \ \ \ \\
}
\and
\IEEEauthorblockN{Georgios Kaissis}
\IEEEauthorblockA{\textit{Technical University of Munich} \\
\textit{Helmholtz Zentrum Munich}\\
Munich, Germany \\
}
\and
\IEEEauthorblockN{\ \ \ \ \ \ \ \ \ \ \ \ \ \ \ \ \ \ \ \ \  \ \ \ \ \ }
\IEEEauthorblockA{ \ \ \ \ \ \ \ \\
}
}

\maketitle

\begin{abstract}
Normalization is an important but understudied challenge in privacy-related application domains such as federated learning (FL), differential privacy (DP), and differentially private federated learning (DP-FL). While the unsuitability of batch normalization for these domains has already been shown, the impact of other normalization methods on the performance of federated or differentially private models is not well-known. To address this, we draw a performance comparison among layer normalization (LayerNorm), group normalization (GroupNorm), and the recently proposed kernel normalization (KernelNorm) in FL, DP, and DP-FL settings. Our results indicate LayerNorm and GroupNorm provide no performance gain compared to the baseline (i.e. no normalization) for shallow models in FL and DP. They, on the other hand, considerably enhance the performance of shallow models in DP-FL and deeper models in FL and DP. KernelNorm, moreover, significantly outperforms its competitors in terms of accuracy and convergence rate (or communication efficiency) for both shallow and deeper models in all considered learning environments. Given these key observations, we propose a kernel normalized ResNet architecture called KNResNet-13 for differentially private learning. Using the proposed architecture, we provide new state-of-the-art accuracy values on the CIFAR-10 and Imagenette datasets, when trained from scratch.
\end{abstract}


\section{Introduction}
\label{sec:intorduction}
\emph{Deep convolutional neural networks} (\emph{CNNs}) are popular in a diverse range of image vision tasks including image classification \cite{krizhevsky2012-deep-cnn1}. Deep CNNs rely on large-scale datasets to effectively train the model, which might be difficult to provide in a centralized manner \cite{horvitz2015data-privacy}. This is because datasets are often distributed across different sites such as hospitals, and contain sensitive data which cannot be transferred to a centralized location due to privacy regulations \cite{bu2020dl-privacy1}. Even if such datasets become available, training algorithms can pose privacy risks to the individuals participating in the dataset, leaking privacy-sensitive information through the trained model \cite{nasr2019-privacy-leakage1, shokri2017-privacy-leakage2, yeom2018-privacy-leakage3}.  

\emph{Federated learning} (\emph{FL}) \cite{mcmahan2017federated-learning} addresses the large-scale data availability challenge by enabling clients to jointly train a global model under the coordination of a central server without sharing their private data. \emph{Network communication}, on the other hand, emerges as a new challenge in federated environments, requiring a large number of communication rounds for model convergence, and exchanging a large amount of traffic in each round \cite{li2020fl-challenges}. FL also causes utility (e.g. in terms of accuracy) reduction due to the \textit{Non-IID} (not independent and identically distributed) nature of the data across the clients \cite{hsieh2020fl-non-iid}. Finally, although FL eliminates the requirement of data sharing, it might still lead to privacy leakage, where the private data of the clients can be reconstructed from the model updates shared with the server \cite{melis2019fl-privacy-1, zhu2020fl-privacy-2, usynin2022fl-privacy-3}.

 \emph{Differential privacy} (DP) \cite{Dwork2014dp} copes with the privacy challenge in both centralized and federated environments by injecting random noise into the model gradients to limit the information learnt about a particular sample in the dataset \cite{abadi2016dp-dl}. DP, however, adversely affects the model utility similar to FL because of the injected noise. In general, there is a trade-off between privacy and utility in DP, where stronger privacy leads to lower utility \cite{alvim2011dp-trade-off}.

\emph{Batch normalization} (BatchNorm) \cite{ioffe2015-batch-norm} is the de facto normalization layer in popular deep CNNs such as ResNets \cite{he2016-deep-resnets} and DenseNets \cite{huang2017-densenet}, which remarkably improves the model convergence rate and accuracy in centralized training. BatchNorm, however, is not suitable for FL and DP settings. This is because BatchNorm relies on the IID distribution of feature values in the batch \cite{ioffe2015-batch-norm}, which is not the case in federated settings. Moreover, per-sample gradients are required to be computed in DP that is impossible for batch-normalized CNNs \cite{abadi2016dp-dl}. \emph{Batch-independent} layers such as \emph{layer normalization} (LayerNorm) \cite{ba2016-layer-norm}, \emph{group normalization} (GroupNorm) \cite{wu2018-group-norm}, and the recently proposed \emph{kernel normalization} (KernelNorm) \cite{nasirigerdeh2022kernel-norm} do not suffer from the BatchNorm's limitations, and therefore, are applicable to FL and DP. 

\textbf{Normalization challenge.} Unsuitability of BatchNorm for federated and differentially private learning has presented a real challenge in the corresponding environments. Unlike the other challenges (i.e. utility, network communication, and privacy), the normalization issue has remained understudied in the context of FL and DP. Previous works \cite{hsieh2020fl-non-iid, zhang2020groupnorm-vs-batchnorm} illustrate that GroupNorm outperforms BatchNorm in terms of accuracy in federated settings. Likewise, GroupNorm also delivers higher accuracy than LayerNorm in differentially private learning \cite{de2022dp-deepmind, klause2022scale-norm, remerscheid2022smoothnet}. Additionally, KernelNorm achieves significantly higher accuracy and faster convergence rate compared to LayerNorm and GroupNorm in both FL and DP settings according to the original study \cite{nasirigerdeh2022kernel-norm}.

However, the prior studies have not made a comparison between different normalization layers and the NoNorm (no normalization layer) case in the first place. Moreover, the experimental evaluation regarding FL and DP environments is limited in the original KernelNorm study \cite{nasirigerdeh2022kernel-norm}, focusing on a cross-silo federated setting (few clients with relatively large datasets) \cite{kairouz2019advances} and a shallow model in DP. Finally, the performance comparisons in the previous works do not consider differentially private federated learning (DP-FL) settings. Given that, two fundamental questions arise: (1) \emph{Do LayerNorm, GroupNorm, and KernelNorm also deliver higher performance than NoNorm in FL, DP, and DP-FL environments?}, and (2) \emph{Does KernelNorm still outperform other normalization layers in cross-device FL (many clients with small datasets), in DP-FL, and using deeper models in DP?}

\textbf{Key findings.} We conduct extensive experiments using the VGG-6 \cite{park2021-vgg6},  ResNet-8 \cite{nasirigerdeh2022kernel-norm}, PreactResNet-18 \cite{he2016-preact-resnet}, and DenseNet20$\times$16 \cite{huang2017-densenet} models trained on the CIFAR-10/100 \cite{cifar-dataset} and Imagenette \cite{imagenette} datasets in FL, DP, and DP-FL settings to address those questions. The findings are as follows: 
\begin{enumerate}
    \item LayerNorm and GroupNorm do not necessarily outperform the NoNorm case for shallow models in FL and DP settings. For instance, LayerNorm and GroupNorm provide slightly lower accuracy and communication efficiency than NoNorm in the cross-silo federated setting, where the shallow VGG-6 model is trained on CIFAR-10. Similarly,  LayerNorm and GroupNorm achieve lower accuracy than NoNorm using the shallow ResNet-8 model on CIFAR-10 in DP (Section \ref{sec:evaluation}).
    \item KernelNorm significantly outperforms NoNorm, LayerNorm, and GroupNorm in terms of communication efficiency (convergence rate) and accuracy in both cross-silo and cross-device FL, with both shallow and deeper models in DP, and using shallow models in DP-FL environments (Section \ref{sec:evaluation}). 
\end{enumerate}
    
\textbf{Solution. } Based on our findings, we advocate employing KernelNorm as the effective normalization layer for FL, DP, and DP-FL settings. Given that, we propose a KernelNorm-based ResNet architecture called KNResNet-13, and show it delivers considerably higher accuracy than the state-of-the-art GroupNorm-based architectures on CIFAR-10 and Imagenette in differentially private learning environments (Section \ref{sec:resnet13_kn}).

\textbf{Contributions. } We make the following contributions: (I) we show LayerNorm and GroupNorm do not deliver higher accuracy than NoNorm with shallow models in FL and DP settings, (II) we illustrate the recently proposed KernelNorm layer has a great potential to become the de facto normalization layer in privacy-enhancing/preserving machine learning, and (III) we propose the KNResNet-13 architecture, and provide new state-of-the-art (SOTA) accuracy values on CIFAR-10 and Imagenette using the proposed architecture in DP environments, when trained from scratch.  

\section{Preliminaries}
\label{sec:preliminary}
\textbf{Federated learning (FL).} A federated environment consists of multiple clients as data holders and a central server as coordinator. FL is a privacy-enhancing technique, which enables the clients to train a global model without sharing their private data with a third party. In FL, or more precisely in the \texttt{FederatedAveraging} (\textit{FedAvg}) algorithm \cite{mcmahan2017federated-learning}, the server randomly chooses $K$ clients, and sends them the global model parameters $W^{g}_{i}$ in each communication round $i$. Next, each selected client $j$ trains the global model on its local dataset using \emph{mini-batch gradient descent}, and shares the local model parameters $W^{l}_{i,j}$ with the server. Finally, the server takes the weighted average over the  local parameters from the clients to update the global model:
\begin{equation*}
    W_{i+1}^{g} = \frac{\sum^{K}_{j=1}N_{j} \cdot W_{i,j}^{l}}{\sum^{K}_{j=1}N_{j}},
\end{equation*}
where $N_{j}$ is the number of samples in client $j$.

A \emph{cross-device} federated setting contains a large number of clients such as mobile devices with small datasets \cite{kairouz2019advances}. The server selects a fraction of clients in each round. Moreover, the underlying assumption is that the communication between clients and server is unstable, and the clients might drop out during training.  A \emph{cross-silo} setting, on the other hand, consists of few clients such as hospitals or research institutions with relatively large datasets and stable network connection \cite{kairouz2019advances}. All clients participate in model training in all communication rounds. For more details on federated learning, the readers are referred to \cite{mcmahan2017federated-learning} and \cite{kairouz2019advances}.

\textbf{Differential privacy (DP).} The differential privacy approach provides a theoretical framework and collection of techniques for privacy-preserving data processing and release \cite{Dwork2014dp}. Its guarantees are formulated in an information-theoretic fashion and describe the upper bound on the multiplicative information gain of an adversary observing the output of a computation over a sensitive database. This definition endows DP with a robust theoretical underpinning and ascertains that its guarantees hold in the presence of adversaries with unbounded prior knowledge and under infinite post-processing. Moreover, DP guarantees are \textit{compositional}, meaning that they degrade predictably when a DP system is executed repeatedly on the same database. Formally, a randomised mechanism $\mathcal{M}$ is said to preserve $(\varepsilon, \delta)$-DP if, for all databases $D$ and $D'$ differing in the data of one individual and all measurable subsets $S$ of the range of $\mathcal{M}$, the following inequality holds:
\begin{equation*}
    \mathbb{P}(\mathcal{M}(D) \in S) \leq e^{\varepsilon} \mathbb{P}(\mathcal{M}(D') \in S) + \delta,
\end{equation*}
where $\mathbb{P}$ is the probability of an event, $\varepsilon \geq 0$ and $0 \leq \delta < 1$. Of note, this inequality must hold also if $D$ and $D'$ are swapped. The guarantee is given over the randomness of $\mathcal{M}$. Intuitively, this characterisation implies that the output of the mechanism should not change \textit{too much} when one individual's data is added or removed from a database, or equivalently, the influence of one individual's data on the result of the computation should be small.

The application of DP to the training of neural networks is usually (and in our work) based on the differentially private stochastic gradient descent (DP-SGD) algorithm \cite{abadi2016dp-dl}. Here, the role of the database is played by the individual (per-sample) gradients of the loss function with respect to the parameters. For the DP guarantee to be well-defined, the intermediate layer outputs (activations), leading to the computation of a per-sample gradient, are not allowed to be influenced by more than one sample. Hence, layers like BatchNorm, which normalize the activations of a layer by considering either other samples in the batch or the statistics of previously seen batches, cannot be employed in DP. We refer the readers to \cite{Dwork2014dp, dwork2008dp-survey, abadi2016dp-dl} for more information on differential privacy.

\textbf{Differentially private federated learning (DP-FL).} Although FL enhances data privacy by eliminating the requirement of data sharing, the model parameters shared with the server can still cause privacy leakage. To overcome this problem, the clients can rely on DP to train the global model on their local data, and share differentially private models with the server. This way, the clients can benefit from the guarantees of DP in federated environments.

\textbf{Normalization.} The normalization layers play a crucial role in deep CNNs. They can smoothen the optimization landscape \cite{santurkar2018-why-bn} and effectively address the problem of vanishing gradients \cite{bengio1994-vanishing-grads1}, leading to improved model performance. The normalization layers are different from each other in their normalization unit, which is a subset of elements from the original input that are normalized together with the mean and variance of the unit \cite{nasirigerdeh2022kernel-norm}. Assume that the input is a 4-dimensional tensor with batch, channel, height, and width as dimensions. BatchNorm \cite{ioffe2015-batch-norm} considers all elements in the batch, height, and width dimensions as its normalization unit. LayerNorm \cite{ba2016-layer-norm}, on the other hand, performs normalization across all elements in the channel, height, and width dimensions but separately for each sample in the batch. The normalization unit of GroupNorm \cite{wu2018-group-norm} contains all elements in the height and width dimensions similar to LayerNorm, but a subset of elements (specified by the group size) in the channel dimension. 

BatchNorm, LayerNorm, and GroupNorm are referred to as \emph{global normalization} layers because they consider all elements in the height and width dimensions during normalization \cite{ortiz2020-lcnorm}. There is also a one-to-one correspondence between the input and output elements in the aforementioned layers, implying that they do not modify the input shape \cite{nasirigerdeh2022kernel-norm}.  These layers have \emph{shift} and \emph{scale} as learnable parameters too for ensuring that the distributions of the input and output elements remain similar \cite{ioffe2015-batch-norm}. In contrast to BatchNorm, LayerNorm and GroupNorm are \emph{batch-independent} because they perform normalization separately for each sample in the batch. 

\textbf{KernelNorm} \cite{nasirigerdeh2022kernel-norm} performs normalization along the channel, height, and width dimensions but independently of the batch dimension akin to LayerNorm and GroupNorm. The normalization unit of KernelNorm, however, is a tensor of shape ($c$, $k_{h}$, $k_{w}$), where $c$ is the number of input channels, and ($k_{h}$, $k_{w}$) is the kernel size. Thus, KernelNorm considers \emph{all elements} in the channel dimension but a \emph{subset of elements} specified by the kernel size from the height and width dimensions during normalization. In simple words, KernelNorm is similar to the pooling layers, except that KernelNorm normalizes the elements instead of computing average or maximum, and carries out operation over all channels rather than on a single channel.

Formally, KernelNorm (1) applies dropout to the original normalization unit $U$ to obtain the \emph{dropped-out} unit  $U^{\prime}$, (2) calculates the mean and variance of $U^{\prime}$, and (3) employs the computed mean and variance to normalize $U$:
\small
\begin{equation}
    U^{\prime} = D_{p}(U),
\end{equation}
\begin{equation}
\label{equ:kn-mu-var}
\begin{aligned}
    \mu_{u^{\prime}} = \frac{1}{c \cdot k_{h} \cdot k_{w}} \cdot \sum_{i_{c}=1}^{c} \sum_{i_{h}=1}^{k_{h}} \sum_{i_{w}=1}^{k_{w}} U^{\prime}(i_{c}, i_{h}, i_{w}), \\
    \sigma^{2}_{u^{\prime}} = \frac{1}{c \cdot k_{h} \cdot k_{w}} \cdot \sum_{i_{c}=1}^{c} \sum_{i_{h}=1}^{k_{h}} \sum_{i_{w}=1}^{k_{w}} (U^{\prime}(i_{c}, i_{h}, i_{w}) - \mu_{u^{\prime}})^{2},
\end{aligned}
\end{equation}
\begin{equation}
\label{equ:kn-norm}
    \hat{U} = \frac{U - \mu_{u^{\prime}}}{\sqrt{\sigma^{2}_{u^{\prime}}  + \epsilon}} ,
\end{equation}
\normalsize
where $p$ is the dropout \cite{srivastava2014-dropout} probability, $\mu_{u^{\prime}}$ and $\sigma^{2}_{u^{\prime}}$ are the mean and variance of $U^{\prime}$, respectively, and $\hat{U}$ is the normalized unit. Partially inspired by BatchNorm, KernelNorm introduces a regularizing effect during training through normalizing the elements of the original unit $U$ via the statistics calculated over the dropped-out unit $U^{\prime}$. 

KernelNorm is a \emph{local normalization} layer. Moreover, it has no learnable parameters, and its output might have very different shape than the input. Similar to LayerNorm and GroupNorm, KernelNorm is batch-independent because it performs normalization separately for each sample of the batch. The \emph{kernel normalized convolutional} (\emph{KNConv}) layer \cite{nasirigerdeh2022kernel-norm} is the combination of the KernelNorm and convolutional layer, where the output of the former is given as input to the latter. 

The modern CNNs are batch-normalized, leveraging the BatchNorm and convolutional layers in their architectures. The corresponding layer/group-normalized networks are obtained by simply replacing BatchNorm with LayerNorm/GroupNorm. The kernel-normalized counterparts \cite{nasirigerdeh2022kernel-norm}, on the other hand, employ the KernelNorm and KNConv layers as the main building blocks, while forgoing the BatchNorm layers. For more details on the normalization layers, the readers can see \cite{ioffe2015-batch-norm, ba2016-layer-norm, wu2018-group-norm, nasirigerdeh2022kernel-norm}.

\section{Evaluation}
\label{sec:evaluation}
We conduct extensive experiments to investigate the performance of different batch-independent normalization layers including LayerNorm, GroupNorm, and KernelNorm in the cross-silo and cross-device FL as well as DP and DP-FL environments. In the following, we first provide the description of the datasets, models, and case studies, and then discuss the results and findings.

\subsection{Experimental Setup}
\label{subsec:exp_setup}

\textbf{Datasets.} The CIFAR-10/100 dataset \cite{cifar-dataset} contains $50000$ train and $10000$ test samples of shape $32 \times 32$ from $10$/$100$ classes. The Imagenette dataset ($160$-pixel version) \cite{imagenette} is a subset of Imagenet \cite{deng2009-imagenet}, including $9469$ train and $3925$ validation images from $10$ "easily classified" labels. The feature values are divided by $255$ for KernelNorm based models, whereas they are normalized using the mean and standard deviation of CIFAR-10/100 or ImageNet for NoNorm, LayerNorm, and GroupNorm based counterparts. The samples of Imagenette are resized to $128\times128$.

\textbf{Models.} We adopt the VGG-6 architecture from \cite{park2021-vgg6}, ResNet-8 model from \cite{nasirigerdeh2022kernel-norm}, PreactResNet-18 implementation from  \cite{preact-resnet-pytorch}, and DenseNet-20$\times$16 (depth of 20 and growth rate of 16) implementation from \cite{densenet-pytorch}. In layer/group-normalized networks, BatchNorm is substituted by LayerNorm/GroupNorm. In the NoNorm case, the BatchNorm layers are either removed or replaced with the identity layer. The kernel-normalized counterparts are implemented by removing the BatchNorm layers,  replacing the convolutional layers with KNConv, and inserting a KernelNorm layer before the final average-pooling layer in the ResNet, PreactResNet, and DenseNet models. In FL, the models employ the ReLU activation. In DP, on the other hand, the activation function is Mish \cite{misra2019-mish}, which was successfully used in \cite{klause2022scale-norm} to achieve SOTA accuracy. We implement the models in the PyTorch library (version 1.11) \cite{Paszke2019-pytorch}. 
\begin{table*}[!ht]
\caption{\textbf{Federated learning}: Test accuracy for different normalization layers; NoNorm (no normalization) slightly outperforms LayerNorm and GroupNorm in (a); KernelNorm delivers higher accuracy than the competitors; B: batch size.}
\label{tab:accuracy-bs-fl}
    \begin{minipage}{.48\textwidth}
        \centering
        \subcaption{CIFAR-10-VGG-6 \textbf{(cross-silo FL)}}
        \vskip -0.05in
        \resizebox{\columnwidth}{!}{\begin{tabular}{lcccc}
            \toprule
             B & NoNorm& LayerNorm & GroupNorm & KernelNorm \\
            \midrule
            16  & 80.19$\pm$0.29 & 78.93$\pm$0.43 & 78.63$\pm$0.56 & \textbf{83.64}$\pm$0.41 \\
            64  & 79.23$\pm$0.31 & 78.97$\pm$0.36 & 79.4$\pm$0.38 & \textbf{82.13}$\pm$0.25 \\
            \bottomrule
            \end{tabular}
        }
    \end{minipage}
    \hspace{0.02\textwidth}
    \begin{minipage}{.48\textwidth}
        \centering
        \subcaption{CIFAR-10-VGG-6 \textbf{(cross-device FL)}}
        \vskip -0.05in
        \resizebox{\columnwidth}{!}{\begin{tabular}{lcccc}
            \toprule
             B & NoNorm & LayerNorm & GroupNorm & KernelNorm \\
            \midrule
                16  & 80.95$\pm$0.27 & 81.89$\pm$0.32 & 81.39$\pm$0.47 & \textbf{84.13}$\pm$0.26 \\
                64  & 80.72$\pm$0.06 & 81.43$\pm$0.19 & 81.44$\pm$0.18 & \textbf{83.77}$\pm$0.11 \\
            \bottomrule
            \end{tabular}
        }
    \end{minipage}
    \par
    \vskip 0.15in
    \centering
    \begin{minipage}{.48\textwidth}
        \centering
        \subcaption{CIFAR-100-PreactResNet-18 \textbf{(cross-silo FL)}}
        \vskip -0.05in
        \resizebox{\columnwidth}{!}{\begin{tabular}{lcccc}
            \toprule
             B & NoNorm & LayerNorm & GroupNorm & KernelNorm \\
            \midrule
            16  & 61.89$\pm$0.13 & 68.16$\pm$0.44 & 67.86$\pm$0.1 & \textbf{71.72}$\pm$0.19 \\
            64  & 60.8$\pm$0.33 & 66.9$\pm$0.41 & 66.45$\pm$0.18 & \textbf{71.29}$\pm$0.21 \\
            \bottomrule
            \end{tabular}
        }
    \end{minipage}
    \hspace{0.02\textwidth}
    \begin{minipage}{.48\textwidth}
        \centering
        \subcaption{CIFAR-100-PreactResNet-18 \textbf{(cross-device FL)}}
        \vskip -0.05in
        \resizebox{\columnwidth}{!}{\begin{tabular}{lcccc}
            \toprule
             B & NoNorm& LayerNorm & GroupNorm & KernelNorm \\
            \midrule
                16  & 63.54$\pm$0.22 & 68.05$\pm$0.92 & 68.23$\pm$0.13 & \textbf{71.75}$\pm$0.24 \\
                64  & 63.33$\pm$0.36 & 67.84$\pm$0.43 & 67.47$\pm$0.24 & \textbf{71.99}$\pm$0.09 \\
            \bottomrule
            \end{tabular}
        }
    \end{minipage}
\vskip -0.1in
\end{table*}

\begin{figure*}[!ht]
    \vskip 0.15in
    \centering
    \begin{minipage}{0.55\textwidth}
        \centering
        \includegraphics[width=1.0\textwidth]{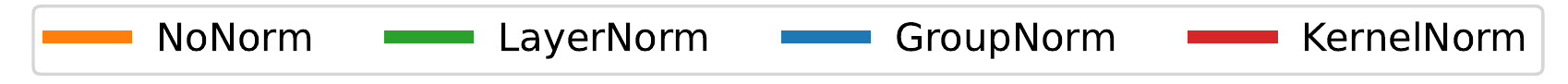}
    \end{minipage}
    \par
	\centering
    \begin{minipage}{0.42\textwidth}
        \centering
        \includegraphics[width=1.0\textwidth]{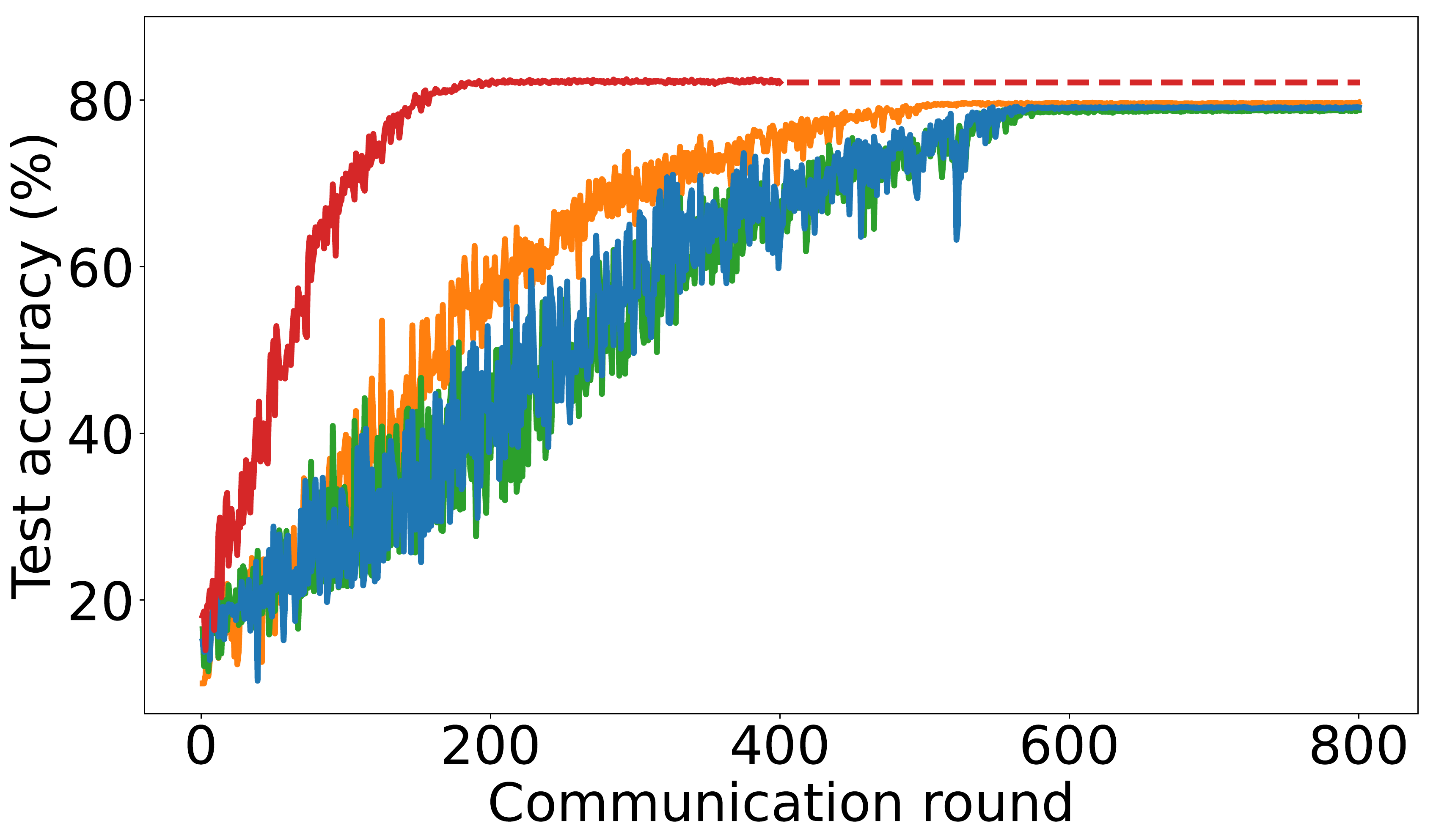}
         \subcaption{CIFAR-10-VGG-6 \textbf{(cross-silo FL)}}
    \end{minipage}
    \hspace{0.02\textwidth}
    \begin{minipage}{0.42\textwidth}
        \centering
        \includegraphics[width=1.0\textwidth]{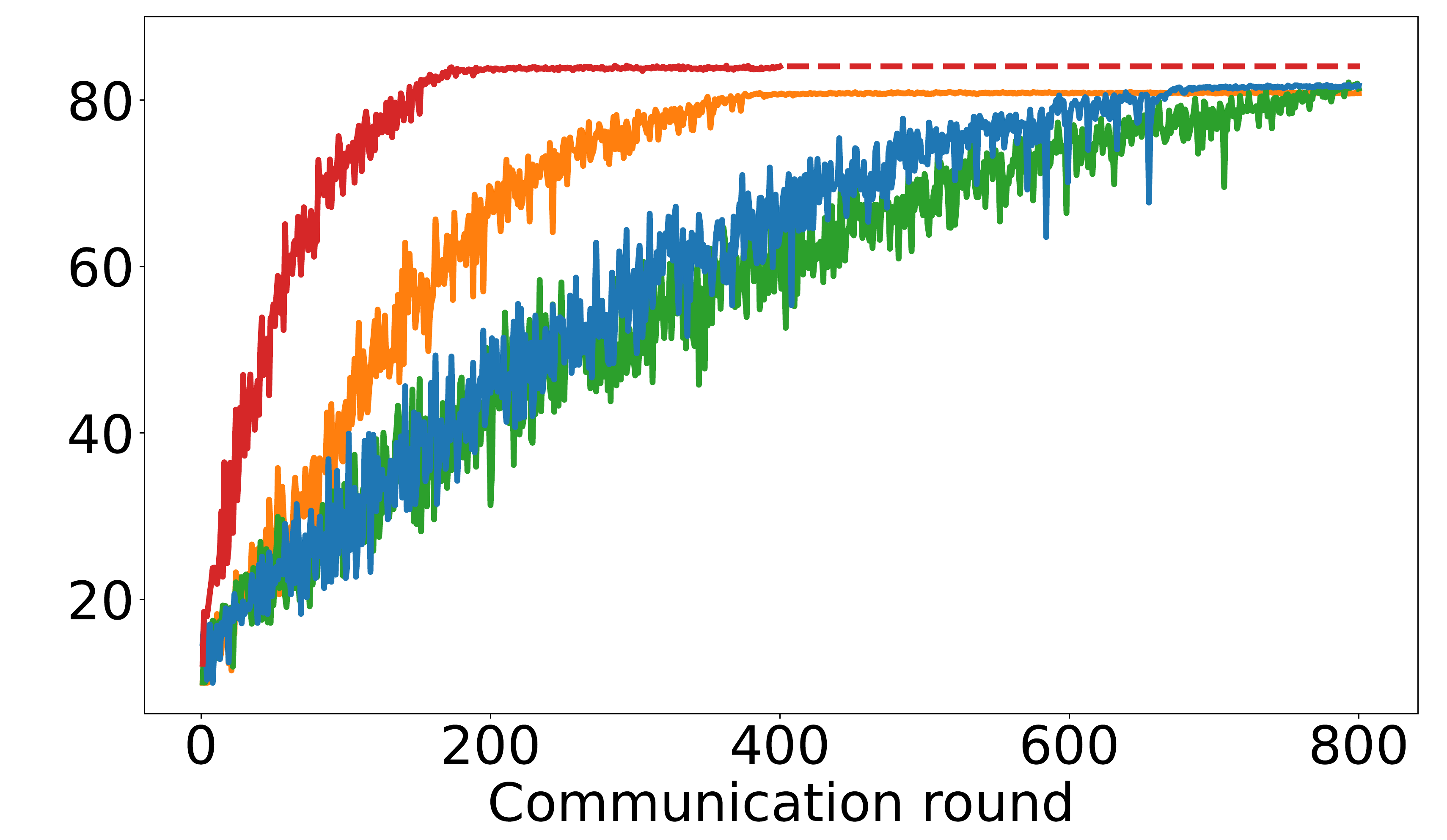}
         \subcaption{CIFAR-10-VGG-6 \textbf{(cross-device FL)}}
    \end{minipage}
    \par
    \vskip 0.2in
	\centering
    \begin{minipage}{0.42\textwidth}
        \centering
        \includegraphics[width=1.0\textwidth]{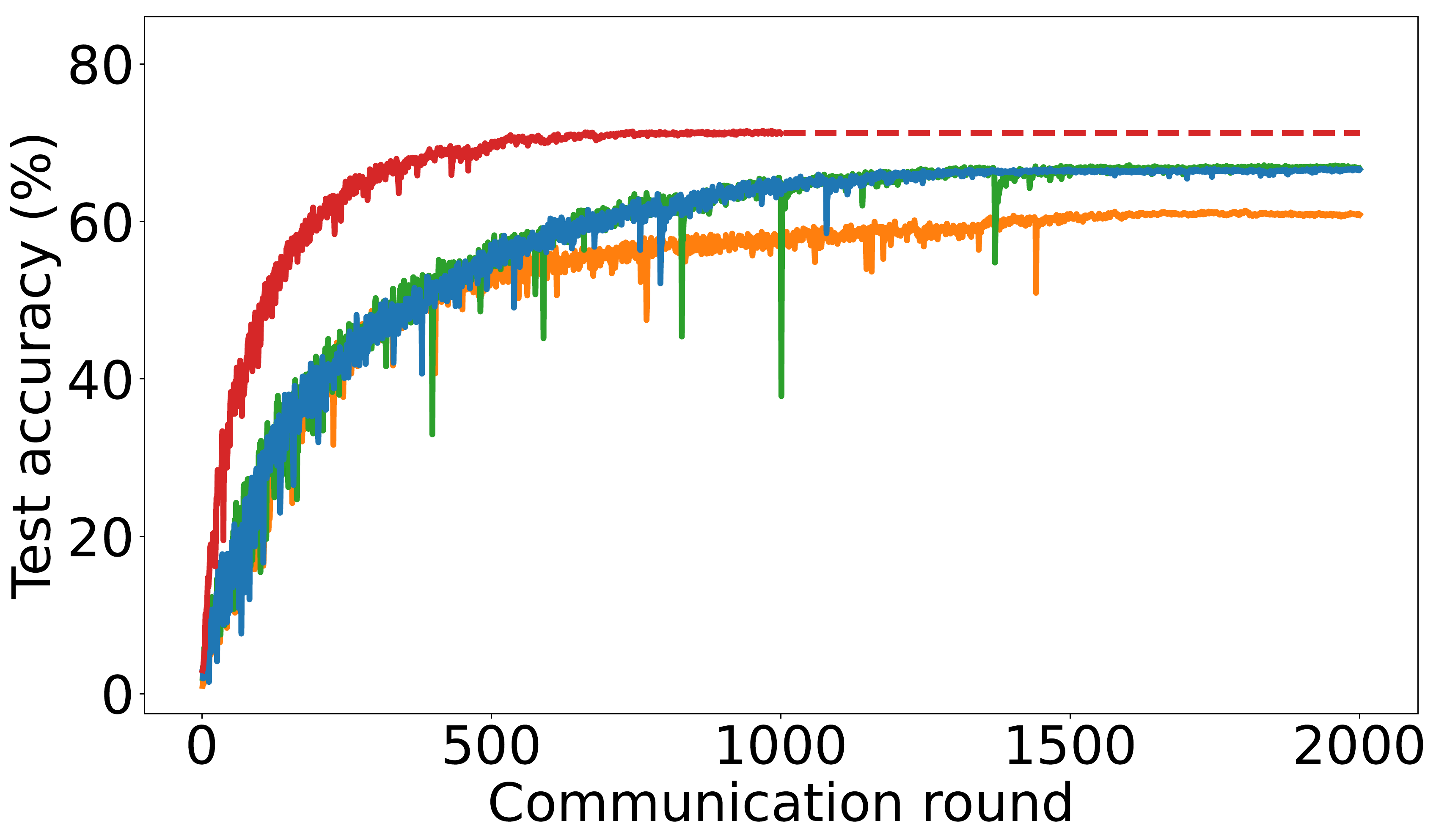}
         \subcaption{CIFAR-100-PreactResNet-18 \textbf{(cross-silo FL)}}
    \end{minipage}
    \hspace{0.02\textwidth}
    \begin{minipage}{0.42\textwidth}
        \centering
        \includegraphics[width=1.0\textwidth]{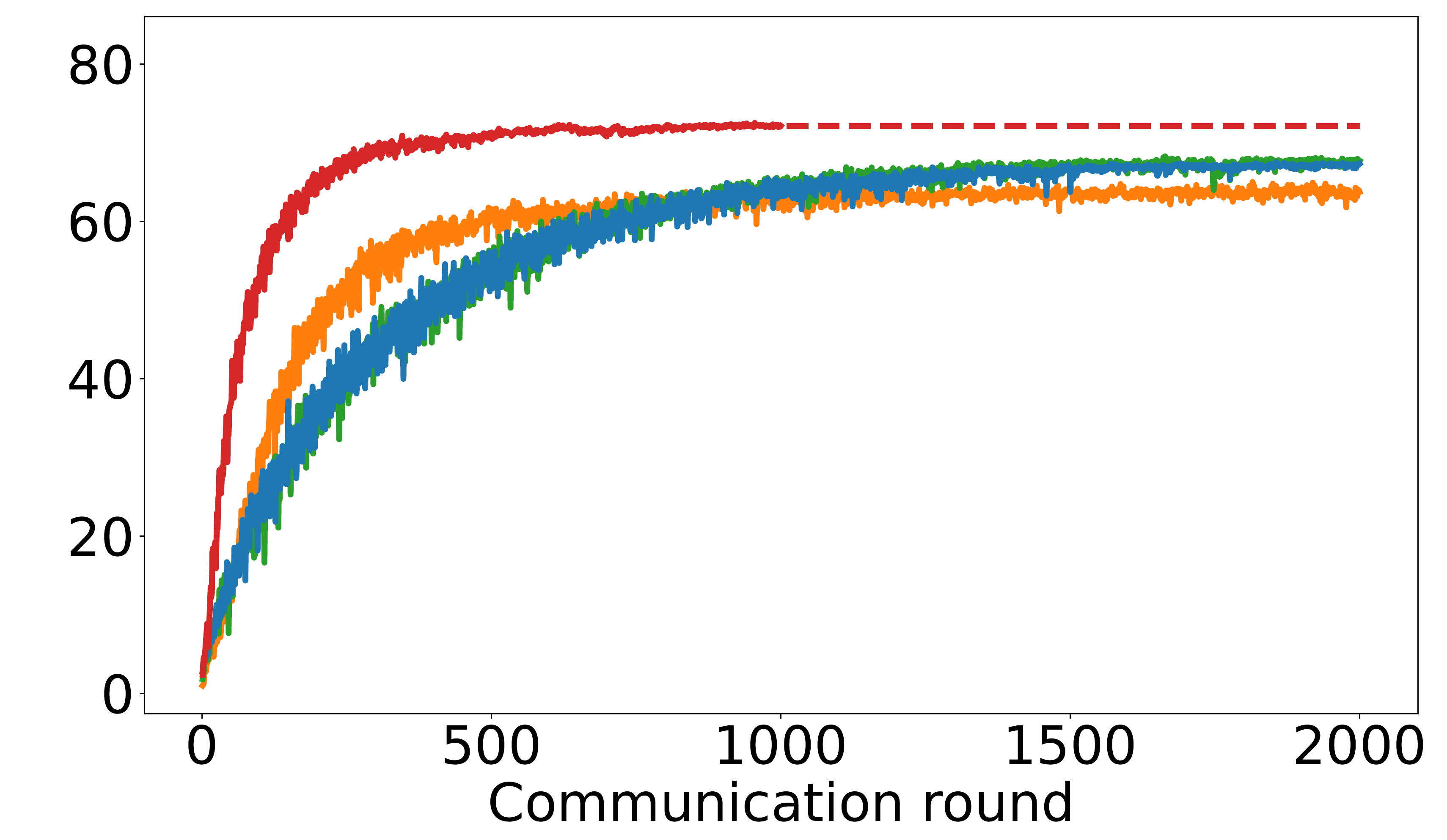}
         \subcaption{CIFAR-100-PreactResNet-18 \textbf{(cross-device FL)}}
    \end{minipage}
    \caption{\textbf{Federated learning:} Communication efficiency for various normalization layers; KernelNorm provides significantly higher communication efficiency than the competitors. Surprisingly, NoNorm outperforms both LayerNorm and GroupNorm in terms of communication efficiency in most cases, i.e (a), (b), (d); batch size is 64.}
    \label{fig:accuracy-round-fl}
\end{figure*}

\textbf{Case Studies.} We design nine different case studies (four in FL, three in DP, and two in DP-FL) to make the performance comparison among the normalization layers:
\begin{enumerate}
    \item \textbf{CIFAR-10-VGG-6 (cross-silo FL):} This case study aims to train the \textit{shallow} VGG-6 model on the \textit{low-resolution} CIFAR-10 dataset in a cross-silo federated environment containing $10$ clients, where each client has samples from only $2$ classes. The sample sizes of the clients are almost the same.
    \item \textbf{CIFAR-10-VGG-6 (cross-device FL):} Similar to the cross-silo counterpart, but in a cross-device federated setting including $100$ clients, where $20$ clients are randomly selected in each round.
    \item \textbf{CIFAR-100-PreactResNet-18 (cross-silo FL):} The aim of this case study is to train the \textit{deeper} PreactResNet-18 model on \textit{more challenging}, low-resolution CIFAR-100 dataset in a cross-silo federated environment consisting of $10$ clients with samples from $20$ labels. The clients have highly similar sample sizes.
    \item \textbf{CIFAR-100-PreactResNet-18 (cross-device FL):} Akin to the cross-silo counterpart, but in a cross-device federated setting consisting of $100$ clients, where $20$ clients are randomly chosen by the server in each round. 
    \item \textbf{CIFAR-10-ResNet-8 (DP):} The goal of this case study is to train the \textit{shallow} ResNet-8 model on the \textit{low-resolution} CIFAR-10 dataset in the DP environment.
    \item \textbf{CIFAR-10-DenseNet-20$\boldmath{\times}$16 (DP):} This case study aims to train the \textit{deeper} DenseNet-20$\times$16 model on the \textit{low-resolution} CIFAR-10 dataset in the DP setting.
    \item \textbf{Imagenette-PreactResNet-18 (DP).} The purpose of this case study is to train the \textit{deeper} PreactResNet-18 model on the \textit{medium-resolution} Imagenette dataset in the differentially private environment.
    \item \textbf{CIFAR-10-VGG-6 (DP-FL):} This case study aims to train the VGG-6 model on the CIFAR-10 dataset in a \textit{differentially private federated setting} with $10$ clients, where the clients have samples from $4$ classes. The sample sizes of the clients are highly similar.
    \item \textbf{CIFAR-10-ResNet-8 (DP-FL):} Similar to the previous case study, but with ResNet-8 as the model.
\end{enumerate}

\textbf{Federated training.} We employ five different values for learning rate tuning in the federated case studies: $\eta$=$\{$0.005, 0.01, 0.025, 0.05, 0.1$\}$. The KernelNorm based models are trained for 400 and 1000 communication rounds in the CIFAR-10 and CIFAR-100 case studies, respectively. The number of rounds for the NoNorm, LayerNorm, and GroupNorm based models is as twice as the kernel normalized counterparts due to their slower convergence rate. The group size is the default value of $32$ for the GroupNorm layer \cite{wu2018-group-norm}. The dropout probability for KNConv and KernelNorm layers are 0.1 and 0.5, respectively. The loss function is cross-entropy, optimizer is SGD with momentum of zero, and training algorithm is FedAvg with number of local epochs of $1$. 

\textbf{Differentially private training.} We set $\varepsilon$=6.0 and $\delta$ = $10^{-5}$ for all DP case studies. Regarding parameter tuning, we use learning rate values of $\eta$=$\{$1.0, 1.5, 2.0$\}$ and clipping values of $C$=$\{$1.0, 1.5, 2.0$\}$. The ResNet-8, DenseNet-20$\times$16, and PreactResNet-18 models are trained for $50$, $70$, and $70$ epochs, respectively. The learning rate is divided by 2 at epochs (T-30) and (T-10), where T is the number of epochs (i.e. 50 or 70). The group size of GroupNorm is 16 for DenseNet-20$\times$16, but 32 for the other models. Notice that we cannot set group size to 32 for DenseNet-20$\times$16 because the number of channels must be divisible by the group size. The dropout probability is 0.1 for all KNConv layers in the kernel normalized models. For ResNet-8, the dropout probability of KernelNorm is 0.25, whereas it is 0.5 for DenseNet-20$\times$16 and PreactResNet-18. 

We employ cross-entropy as loss function, zero-momentum SGD as optimizer, and the Opacus library (version 1.1) \cite{opacus} for model training. We observe that changing the kernel size of the shortcut connections in PreactResNet-18 from $1$$\times$$1$ to $2$$\times$$2$ slightly enhances the accuracy of the kernel normalized model, but provides no accuracy gain for the competitors. Thus, the aforementioned kernel size remains $1$$\times$$1$ for NoNorm, LayerNorm, and GroupNorm, whereas it is $2$$\times$$2$ for KernelNorm.

\textbf{Differentially private federated training.} We set $\varepsilon$=8.0 and $\delta$=$10^{-5}$ for both DP-FL case studies. We leverage learning rate values of $\eta$=$\{$0.01, 0.025, 0.05$\}$ and clipping values of $C$=$\{$1.0, 1.5, 2.0$\}$ for parameter tuning. The group size of GroupNorm is $32$, and the dropout probabilities of the KNConv and KernelNorm layers are 0.1 and 0.25, respectively. The models are trained for $100$ communication rounds with a fixed learning rate. The loss function, optimizer, and training algorithm are cross-entropy, SGD with momentum of zero, and FedAvg with number of local epochs of 1, respectively.

\subsection{Results}
\label{subsec:result}
For all case studies, we first determine the optimal learning rate (and clipping value) based on the model accuracy on the test dataset (see Appendix).  We repeat the experiment achieving the highest accuracy three times and report mean/median/mean and the standard deviation of the runs for the FL/DP/DP-FL case studies. We consider the average over the last 10 communication rounds, final accuracy, and the average over the last 3 rounds as the representative accuracy of the run in the FL, DP, and DP-FL settings, respectively. 
\begin{table*}[!ht]
\caption{\textbf{Differential privacy}: Test accuracy for various normalization layers; NoNorm (no normalization) delivers slightly higher accuracy than LayerNorm and GroupNorm in (a); KernelNorm considerably outperforms the competitors; $\varepsilon$=$6.0$, $\delta$=$10^{-5}$.}
\vskip 0.05in
\label{tab:accuracy-bs-dp}
    \begin{minipage}{.48\textwidth}
        \centering
        \subcaption{CIFAR-10-ResNet-8 (DP)}
        \vskip -0.05in
        \resizebox{\columnwidth}{!}{\begin{tabular}{lllll}
            \toprule
             B & NoNorm& LayerNorm & GroupNorm & KernelNorm \\
            \midrule
                512 & 65.11$\pm$0.29& 70.01$\pm$0.19& 70.27$\pm$0.08& \textbf{72.18}$\pm$0.15 \\
                1024 & 69.05$\pm$0.4& 71.38$\pm$0.5& 71.75$\pm$0.45& \textbf{74.31}$\pm$0.14 \\
                2048 & 72.7$\pm$0.25& 71.67$\pm$0.42& 71.73$\pm$0.31& \textbf{75.46}$\pm$0.34 \\
                3072 & 71.99$\pm$0.14& 69.39$\pm$0.27& 68.99$\pm$0.27& \textbf{75.48}$\pm$0.24 \\
            \bottomrule
            \end{tabular}
        }
    \end{minipage}
    \hspace{0.02\textwidth}
    \begin{minipage}{.48\textwidth}
        \centering
        \subcaption{CIFAR-10-DenseNet-20$\times$16 (DP)}
        \vskip -0.05in
        \resizebox{\columnwidth}{!}{\begin{tabular}{lcccc}
            \toprule
             B & NoNorm & LayerNorm & GroupNorm & KernelNorm \\
            \midrule
            256 & 57.03$\pm$0.48& 65.62$\pm$0.7& 66.16$\pm$0.56& \textbf{68.49}$\pm$0.24 \\
            512 & 64.15$\pm$0.74& 69.24$\pm$0.68& 68.72$\pm$0.65& \textbf{70.86}$\pm$0.44 \\
            1024 & 64.98$\pm$0.6& 69.68$\pm$0.8& 69.57$\pm$0.97& \textbf{72.74}$\pm$0.34 \\
            2048 & 65.29$\pm$0.53& 66.66$\pm$0.78& 67.31$\pm$0.26& \textbf{72.49}$\pm$0.39 \\
            \bottomrule
            \end{tabular}
        }
    \end{minipage}
    \par
    \vskip 0.15in
    \centering
    \begin{minipage}{.48\textwidth}
        \centering
        \subcaption{Imagenette-PreactResNet-18 (DP)}
        \vskip -0.05in
        \resizebox{\columnwidth}{!}{\begin{tabular}{lcccc}
            \toprule
             B & NoNorm& LayerNorm & GroupNorm & KernelNorm \\
            \midrule
                512 & 25.27$\pm$3.95& 54.83$\pm$0.65& 56.7$\pm$0.19& \textbf{59.1}$\pm$0.33 \\
                1024 & 53.69$\pm$0.83& 54.54$\pm$0.23& 57.17$\pm$0.42& \textbf{58.9}$\pm$0.42 \\
                2048 & 53.53$\pm$0.99& 53.3$\pm$0.32& 54.59$\pm$0.27& \textbf{56.11}$\pm$0.26 \\
            \bottomrule
            \end{tabular}
        }
    \end{minipage}
\vskip -0.1in
\end{table*}

\begin{figure*}[!ht]
    \vskip 0.02 in
    \centering
    \begin{minipage}{0.55\textwidth}
        \centering
        \includegraphics[width=1.0\textwidth]{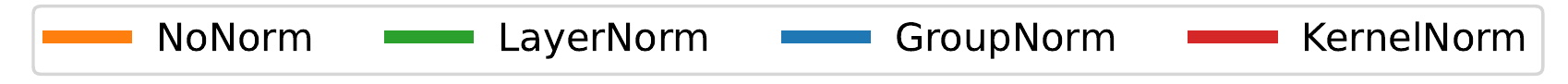}
    \end{minipage}
    \par
	\centering
    \begin{minipage}{0.42\textwidth}
        \centering
        \includegraphics[width=1.0\textwidth]{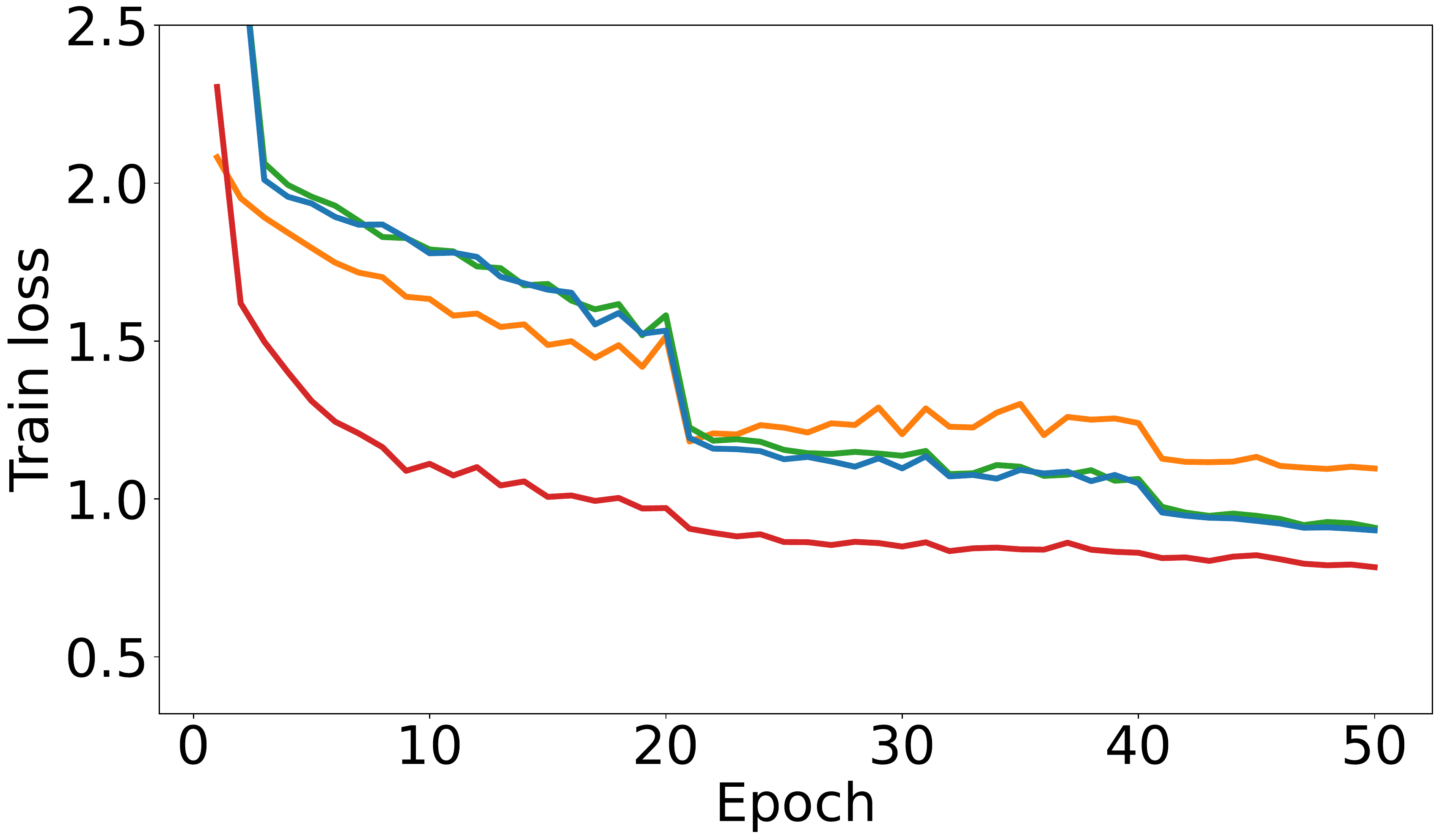}
    \end{minipage}
    \hspace{0.02\textwidth}
    \begin{minipage}{0.42\textwidth}
        \centering
        \includegraphics[width=1.0\textwidth]{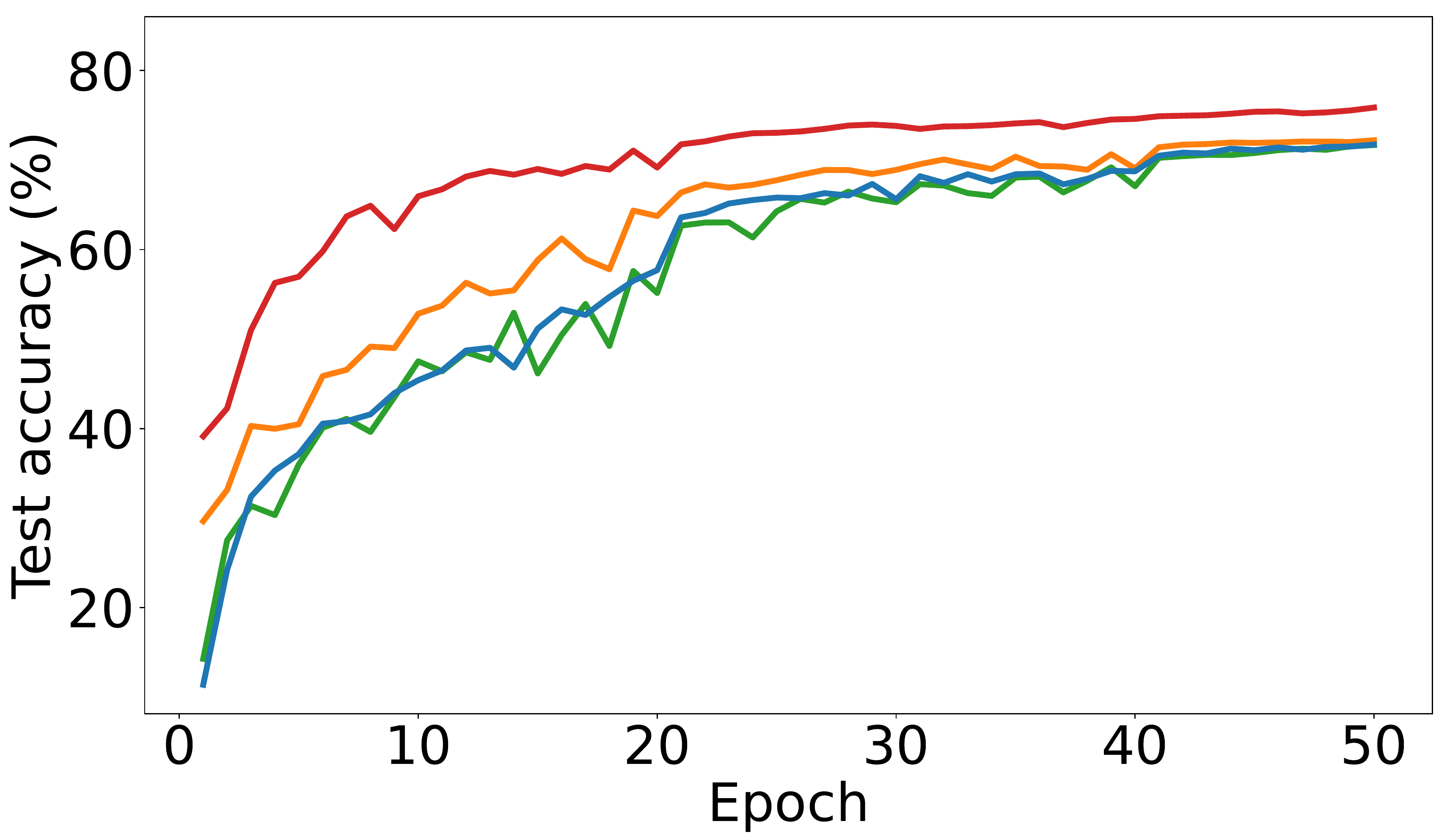}
    \end{minipage}
    \vskip -0.1in
    \begingroup
        \subcaption{CIFAR-10-ResNet-8 (DP)}
    \endgroup
    
    \vskip 0.15in
    \centering
    \begin{minipage}{0.42\textwidth}
        \centering
        \includegraphics[width=1.0\textwidth]{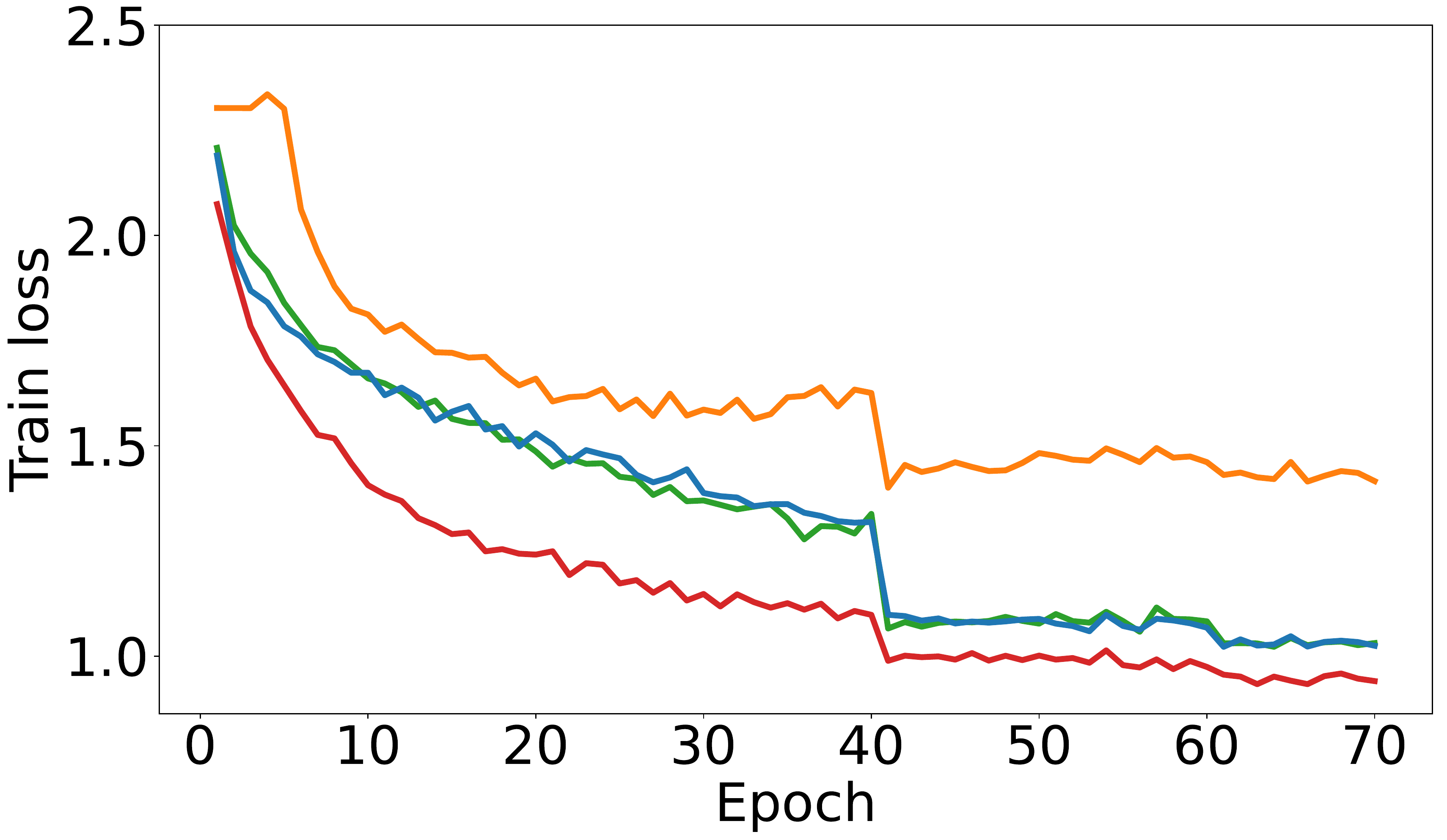}
    \end{minipage}
    \hspace{0.02\textwidth}
    \begin{minipage}{0.42\textwidth}
        \centering
        \includegraphics[width=1.0\textwidth]{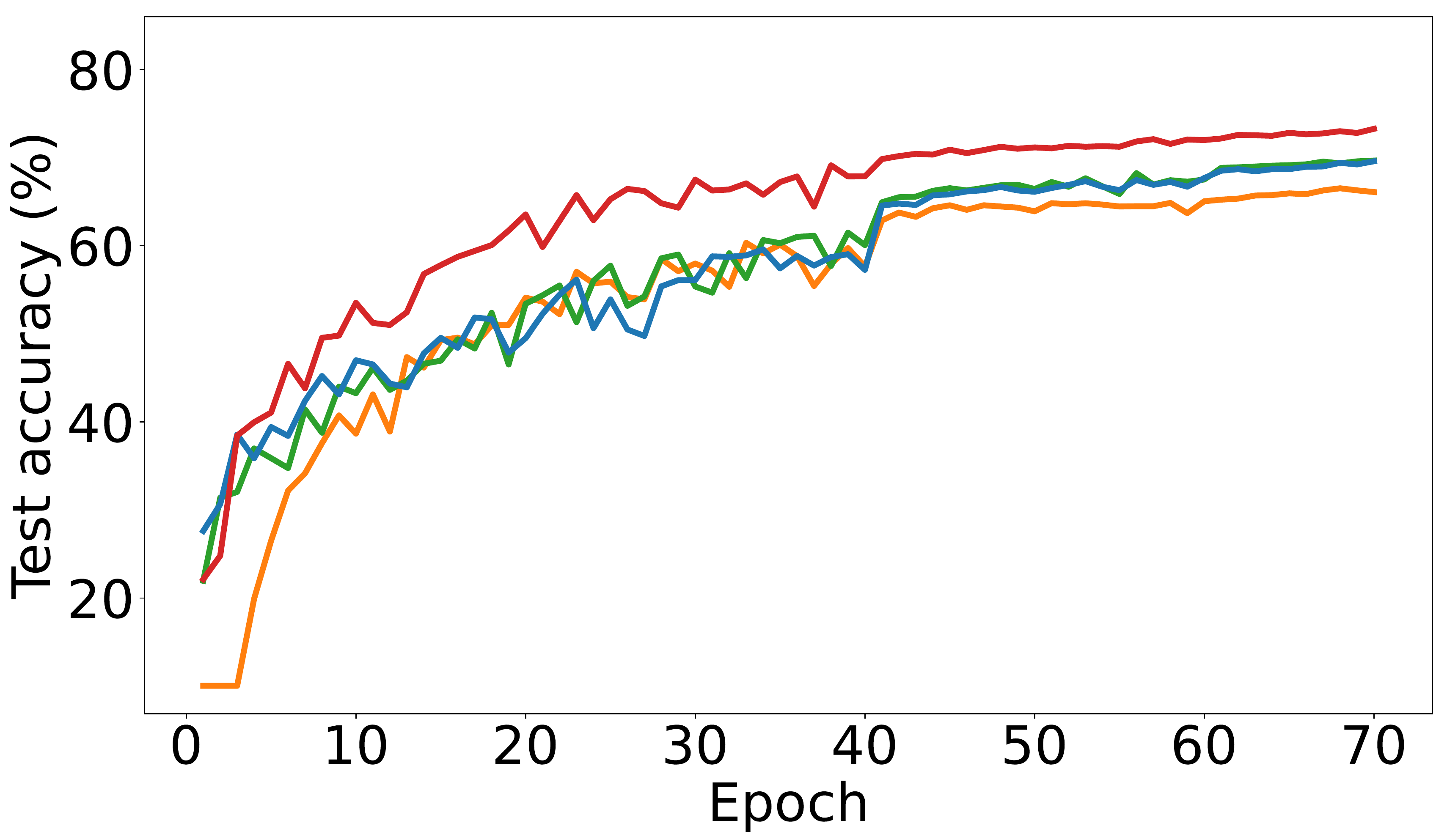}
    \end{minipage}
    \vskip -0.1in
    \begingroup
        \subcaption{CIFAR-10-DenseNet-20$\times$16 (DP)}
    \endgroup
    
    \vskip 0.15in
    \centering
    \begin{minipage}{0.42\textwidth}
        \centering
        \includegraphics[width=1.0\textwidth]{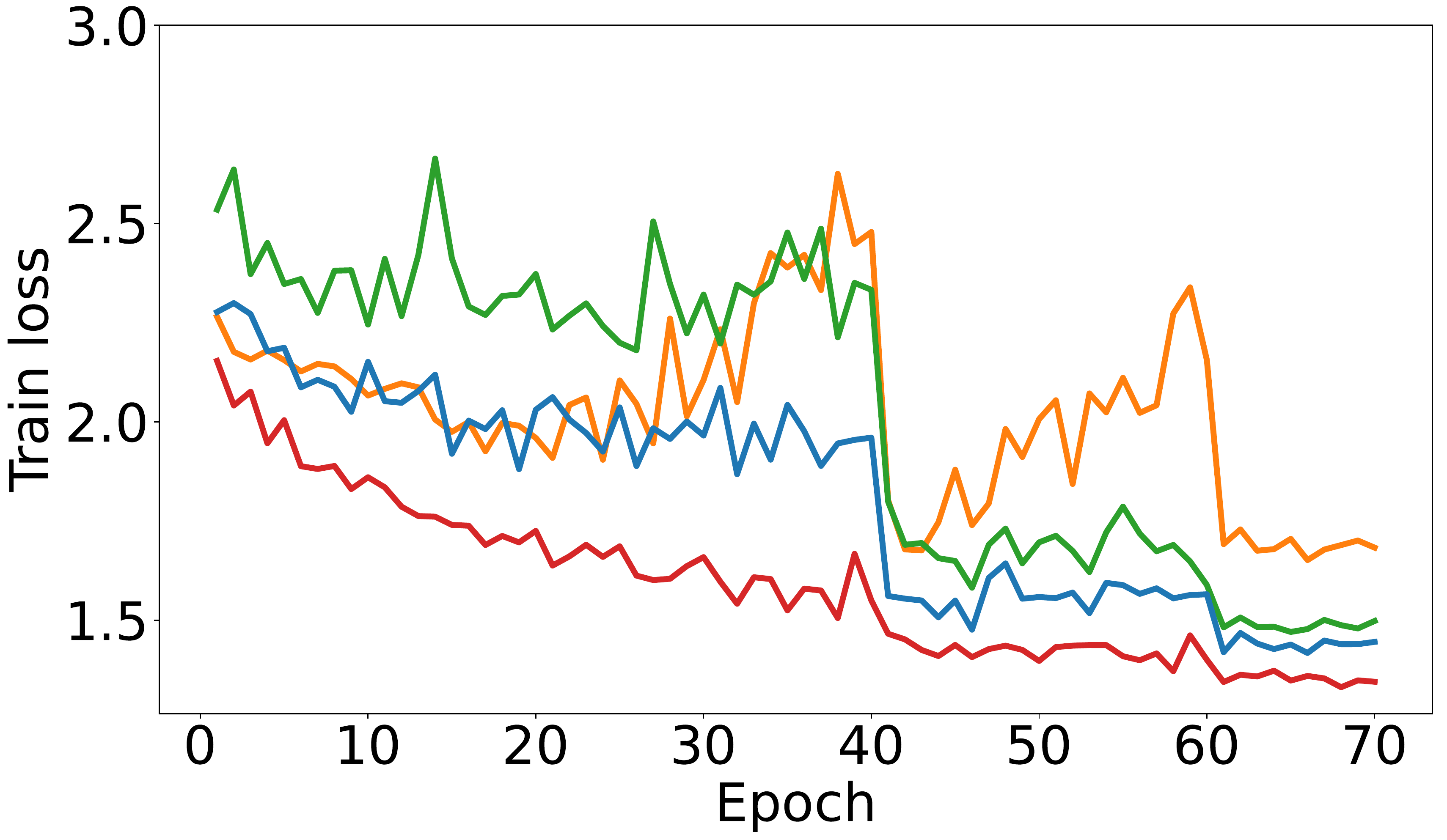}
    \end{minipage}
    \hspace{0.02\textwidth}
    \begin{minipage}{0.42\textwidth}
        \centering
        \includegraphics[width=1.0\textwidth]{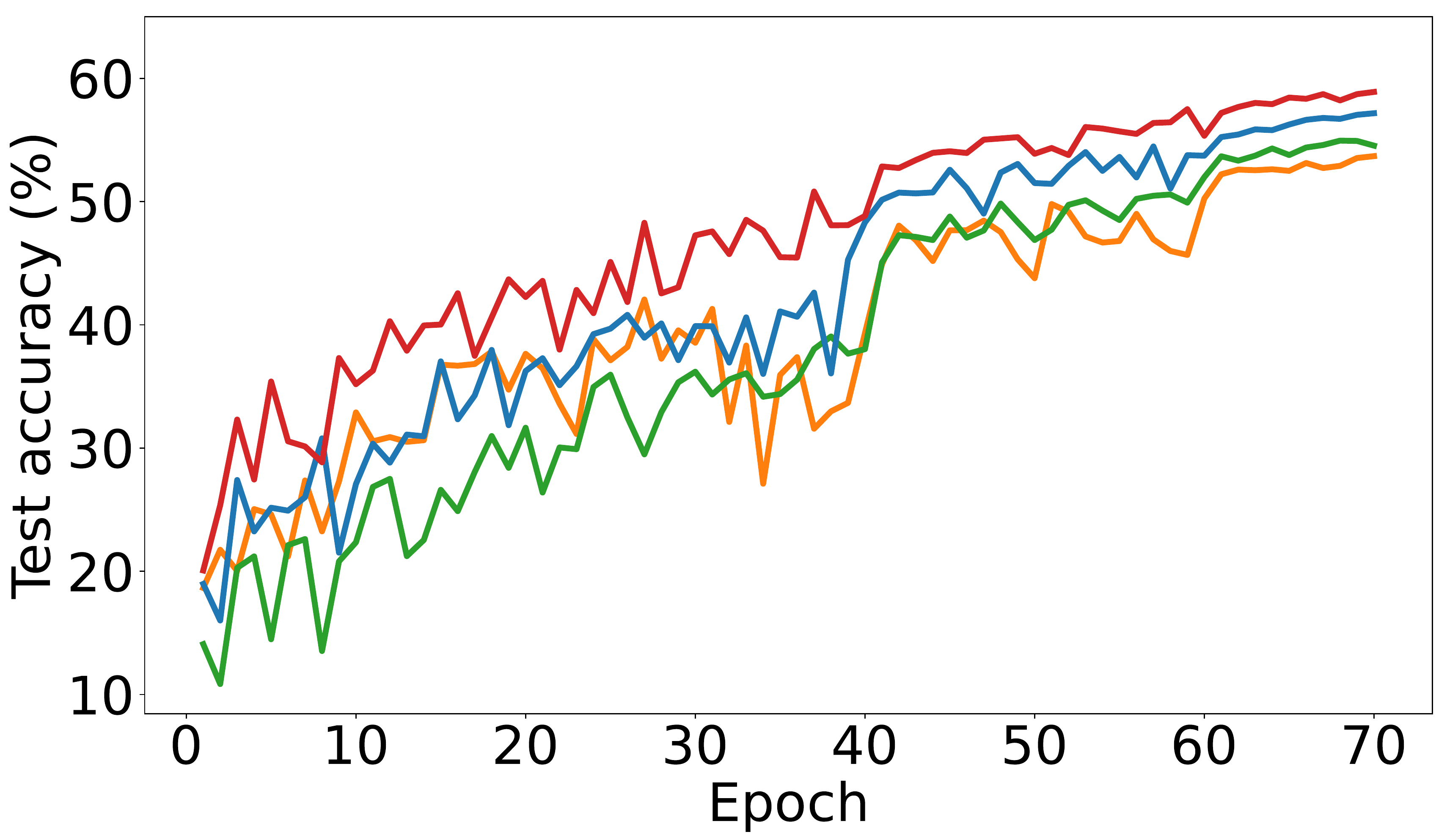}
    \end{minipage}
    \vskip -0.1in
    \begingroup
        \subcaption{Imagenette-PreactResNet-18 (DP)}
    \endgroup
    
    \caption{\textbf{Differential privacy}: Convergence rate for different normalization layers; kernel normalized models provides much faster convergence rate than the competitors; batch size is $2048$, $1024$, and $1024$ for (a), (b), and (c), respectively.}
    \label{fig:accuracy-epoch-dp}
\end{figure*}

\begin{table*}[!ht]
\caption{ 
\textbf{Differentially private federated learning}: Test accuracy for different normalization layers; KernelNorm delivers considerably higher accuracy than the competitors; $\varepsilon$=$8.0$, $\delta$=$10^{-5}$.
}
\vskip 0.05in
\label{tab:accuracy-bs-dp-fl}
    \begin{minipage}{.48\textwidth}
        \centering
        \subcaption{CIFAR-10-VGG-6 (DP-FL)}
        \vskip -0.05in
        \resizebox{\columnwidth}{!}{\begin{tabular}{lllll}
            \toprule
             B & NoNorm& LayerNorm & GroupNorm & KernelNorm \\
            \midrule
            256  & 30.5$\pm$0.44 & 38.23$\pm$0.37 & 37.29$\pm$0.71 & \textbf{46.79}$\pm$0.81 \\
            512  & 29.73$\pm$1.01 & 39.47$\pm$0.48 & 39.75$\pm$0.65 & \textbf{45.37}$\pm$0.22 \\
            1024  & 33.43$\pm$1.33 & 39.19$\pm$0.64 & 38.85$\pm$0.97 & \textbf{47.11}$\pm$0.37 \\
            \bottomrule
            \end{tabular}
        }
    \end{minipage}
    \hspace{0.02\textwidth}
    \begin{minipage}{.48\textwidth}
        \centering
        \subcaption{CIFAR-10-ResNet-8 (DP-FL)}
        \vskip -0.05in
        \resizebox{\columnwidth}{!}{\begin{tabular}{lcccc}
            \toprule
             B & NoNorm & LayerNorm & GroupNorm & KernelNorm \\
            \midrule
                256  & 34.76$\pm$0.95 & 38.43$\pm$1.48 & 40.69$\pm$1.03 & \textbf{45.18}$\pm$0.34 \\
                512  & 36.11$\pm$0.7 & 41.09$\pm$0.33 & 41.8$\pm$0.41 & \textbf{46.75}$\pm$0.48 \\
                1024  & 38.19$\pm$0.19 & 41.41$\pm$1.08 & 41.39$\pm$0.82 & \textbf{48.45}$\pm$1.09 \\
            \bottomrule
            \end{tabular}
        }
    \end{minipage}
\vskip -0.1in
\end{table*}

\begin{figure*}[!ht]
    \centering
    \begin{minipage}{0.55\textwidth}
        \centering
        \includegraphics[width=1.0\textwidth]{imgs/fl/legend.pdf}
    \end{minipage}
    \par
	\centering
    \begin{minipage}{0.42\textwidth}
            \centering
             \includegraphics[width=1.0\textwidth]{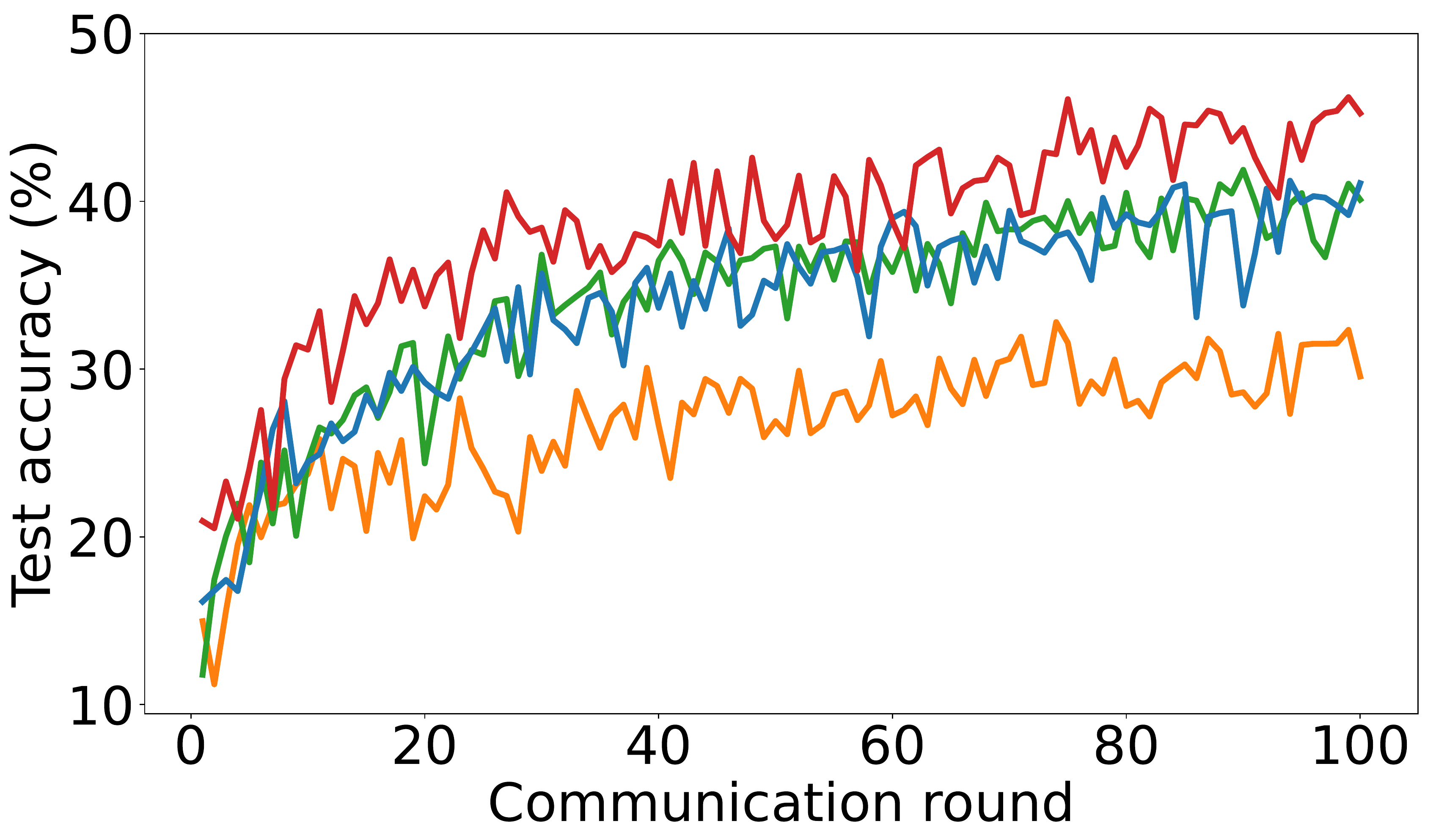}
             \subcaption{CIFAR-10-VGG-6 (DP-FL)}
    \end{minipage}
        \hspace{0.02\textwidth}
    \begin{minipage}{0.42\textwidth}
            \centering
            \includegraphics[width=1.0\textwidth]{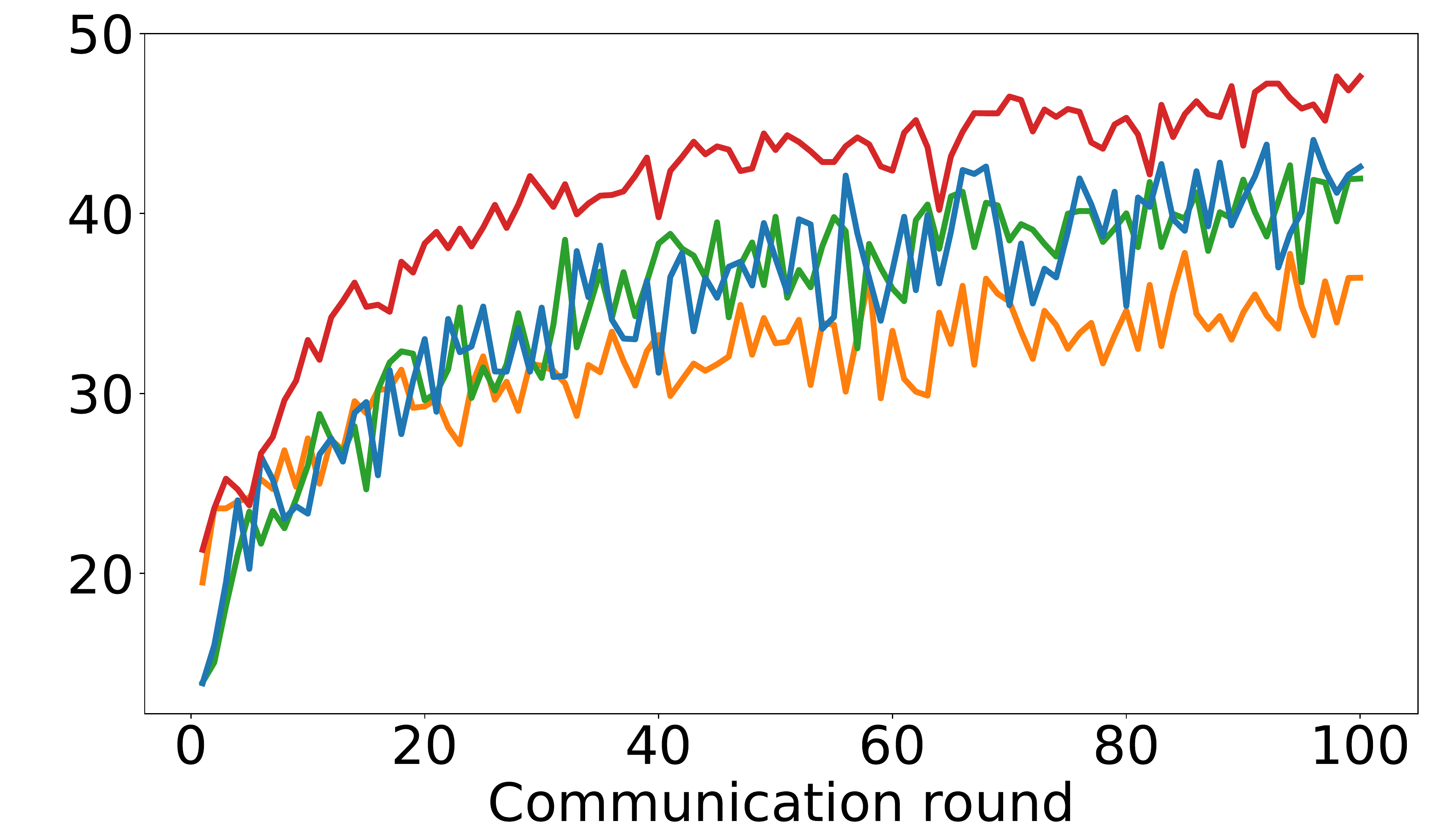}
             \subcaption{CIFAR-10-ResNet-8 (DP-FL)}
    \end{minipage}
    
    \caption{
    \textbf{Differentially private federated learning}: Convergence rate for various normalization layers; kernel normalized models deliver higher convergence rate than the competitors; batch size is $512$.}
    
    \label{fig:accuracy-round-dp-fl}
\end{figure*}

\textbf{Federated learning.} Table \ref{tab:accuracy-bs-fl} lists the test accuracy values for the FL case studies. According to the table, (1) NoNorm slightly outperforms LayerNorm and GroupNorm in the CIFAR-10-VGG-6 (cross-silo FL) case study, whereas LayerNorm and GroupNorm deliver higher accuracy compared to NoNorm in the other case studies; (2) KernelNorm achieves considerably higher accuracy than the competitors. Fig. \ref{fig:accuracy-round-fl} illustrates the communication efficiency (i.e. accuracy versus communication round) for the FL case studies. As shown in the figure, (1)  NoNorm, surprisingly, provides higher communication efficiency than LayerNorm and GroupNorm for most case studies; (2) KernelNorm achieves remarkably higher communication efficiency compared with NoNorm, LayerNorm, and GroupNorm.

\textbf{Differential privacy.} Table \ref{tab:accuracy-bs-dp} and Fig. \ref{fig:accuracy-epoch-dp} demonstrate the test accuracy and convergence rate of different normalization layers for the DP case studies, respectively. According to the table and figure, (1) NoNorm slightly outperforms LayerNorm and GroupNorm in terms of accuracy in the  CIFAR-10-ResNet-8 (DP) case study, but LayerNorm and GroupNorm achieve higher accuracy compared to NoNorm in the other case studies, (2) KernelNorm provides higher accuracy than the competitors in all DP case studies, and (3) KernelNorm based models converge much faster than those based on NoNorm, LayerNorm, and GroupNorm. 

\textbf{Differentially private federated learning.} Table \ref{tab:accuracy-bs-dp-fl} lists the test accuracy values, and Fig. \ref{fig:accuracy-round-dp-fl} illustrates the convergence rate of different normalization layers for the DP-FL case studies. As shown in the table and figure, (1) the NoNorm based models deliver much lower accuracy and slower convergence rate than  LayerNorm, GroupNorm, and KernelNorm based ones, and (2) the kernel normalized models achieve considerably higher accuracy and faster convergence rate than the competitors.

\subsection{Findings}
\label{subsec:findings}
Based on our experimental evaluation, (I) LayerNorm and GroupNorm do not necessarily outperform NoNorm in shallow networks such as VGG-6/ResNet-8 under the FL/DP settings. However, they achieve significant accuracy gain compared to NoNorm for deeper models  (e.g. DenseNet-20$\times$16 and PreactResNet-18) in FL and DP as well as shallow models in DP-FL, and (II) KernelNorm delivers remarkably higher accuracy and convergence rate (communication efficiency) than NoNorm, LayerNorm, and GroupNorm with both shallow and deeper networks trained in FL (cross-silo and cross-device) and DP as well as shallow models in DP-FL. Therefore, KernelNorm is the most effective normalization method for FL, DP, and DP-FL settings.

\section{Kernel Normalized ResNet-13}
\label{sec:resnet13_kn}
The experimental results from the previous section indicate KernelNorm outperforms the competitors in the DP setting using models that originally designed based on global normalization layers such as BatchNorm (e.g. PreactResNets or DenseNets). The existing architectures, however, are not necessarily optimal for KernelNorm. For instance, the kernel size of $1$$\times$$1$ in the shortcut connections of the ResNet architecture is not beneficial for KernelNorm, which requires kernel sizes greater than $1$ to benefit from the spatial correlation of the elements during normalization. 

\begin{figure*}[!ht]
	\centering
    \begin{minipage}{0.42\textwidth}
        \centering
        \includegraphics[width=1.0\textwidth]{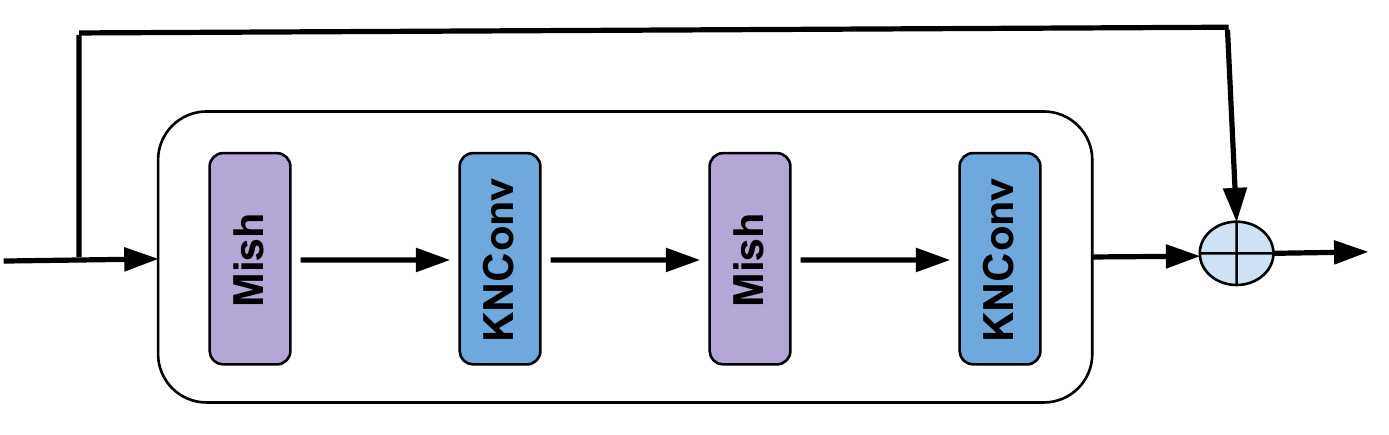}
         \subcaption{Residual block}
    \end{minipage}
    \hspace{0.1\textwidth}
    \begin{minipage}{0.3\textwidth}
        \centering
        \includegraphics[width=1.0\textwidth]{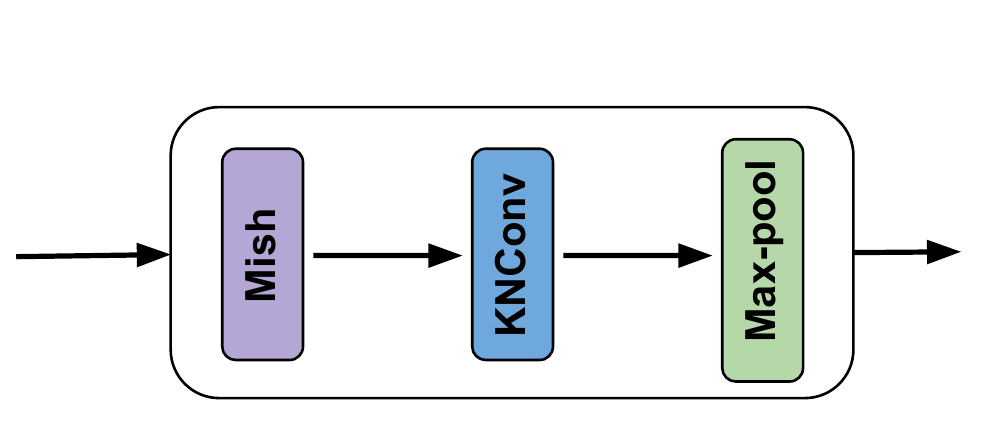}
         \subcaption{Transitional block}
    \end{minipage}
    \par
    \vskip 0.2in
	\centering
    \begin{minipage}{0.8\textwidth}
        \centering
        \includegraphics[width=1.0\textwidth]{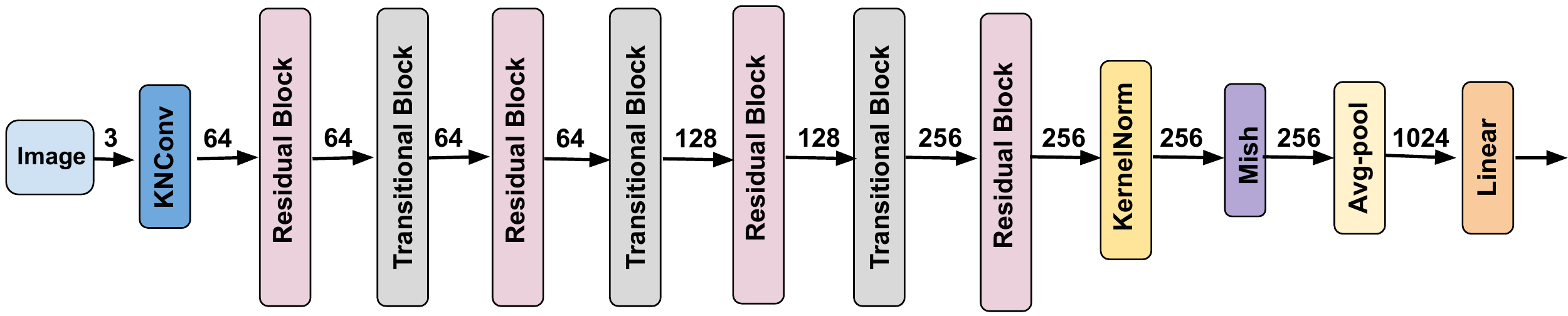}
         \subcaption{KNResNet-13 architecture}
    \end{minipage}
    \caption{\textbf{KNResNet-13 architecture} consists of kernel normalized residual and transitional blocks. The kernel size, stride, and padding of the KNConv layers are $3 \times 3$, $1 \times 1$, and $1 \times 1$, respectively. The kernel size of max-pooling is $2 \times 2$. The dropout probability of KNConv and KernelNorm are $0.1$ and $0.5$, respectively. For medium-resolution images, the first KNConv layer is replaced by a KNConv layer with kernel size $7 \times 7$, stride $2 \times 2$, and padding $3 \times 3$, followed by a Mish activation and $2 \times 2$ max-pooling layer. The numbers indicate the input/output channels (filters) of KNConv or neurons of the linear layer.}
    \label{fig:knresnet13}
\end{figure*}

Given that, we propose a bespoke ResNet architecture for KernelNorm (Fig. \ref{fig:knresnet13}) to improve the SOTA accuracy values on the CIFAR-10 and Imagenette datasets in differentially private learning settings. We refer to the proposed architecture as \textit{KNResNet-13}, which includes twelve kernel normalized convolutional layers and a final classification (linear) layer. 

The convolutional blocks in KNResNet-13 are either residual (Fig. \ref{fig:knresnet13}a) or transitional (Fig. \ref{fig:knresnet13}b). The residual blocks contain two KNConv layers with the same number of input and output channels. The transitional blocks include a KNConv and max-pooling layer, aiming to downsample the input. All KNConv layers have kernel size $3\times3$, stride $1\times1$, padding $1\times1$, and dropout probability $0.1$. The kernel size of the max-pooling layers is $2 \times 2$. The architecture employs Mish as the activation function. The last residual block is followed by a KernelNorm layer with dropout probability $0.5$, Mish activation, $2 \times 2$ adaptive average-pooling, and linear layer with $1024$ neurons. For medium-resolution images (e.g. $224 \times 224$), the first KNConv layer is replaced by a $7 \times 7$ KNConv layer followed by the Mish activation and $2 \times 2$ max-pooling layer. 

\begin{table*}[!ht]
\caption{\textbf{Differential privacy}: Comparison of the test accuracy values from the proposed KNResNet-13 architecture with those from the recent studies; $\delta$=$10^{-5}$.}
\vskip 0.025in
\label{tab:accuracy-comparision}
    \centering
    \begin{minipage}{.8\textwidth}
        \subcaption{CIFAR-10}
        \vskip -0.05in
        \resizebox{\columnwidth}{!}{\begin{tabular}{lllll}
            \toprule
             Study & Model & Normalization & $\varepsilon$ & Test accuracy \\
            \midrule
                \textit{Klause et al.} (2022) \cite{klause2022scale-norm} & ResNet-9 & GroupNorm & 9.88 & 73.0 \\
                \textit{Nasirigerdeh et al.} (2022) \cite{nasirigerdeh2022kernel-norm} & ResNet-8 & KernelNorm & 8.0 & 76.66 \\
                \textit{Ours} & KNResNet-13 & KernelNorm & 8.0 & \textbf{78.51}$\pm$0.35\\
            \midrule
                \textit{D{\"o}rmann et al.} (2021) \cite{dormann2021dp-comparison1} & VGG-8 & NoNorm & 7.42 & 70.1 \\
                \textit{Klause et al.} (2022) \cite{klause2022scale-norm} & ResNet-9 & GroupNorm & 7.42 & 71.8 \\
                \textit{Remerscheid et al.} (2022) \cite{remerscheid2022smoothnet} & DenseNet-14 & GroupNorm & 7.0 & 73.5 \\
                \textit{Nasirigerdeh et al.} (2022) & ResNet-8 & KernelNorm & 6.0 & 75.46 \\
                \textit{Ours} & KNResNet-13 & KernelNorm & 6.0 & \textbf{77.09}$\pm$0.31\\
            \midrule
                \textit{D{\"o}rmann et al.} (2021) \cite{dormann2021dp-comparison1} & VGG-8 & NoNorm & 4.21 & 66.2 \\
                \textit{Nasirigerdeh et al.} (2022) & ResNet-8 & KernelNorm & 4.0 & 73.32 \\
                \textit{Ours} & KNResNet-13 & KernelNorm & 4.0 & \textbf{74.51}$\pm$0.19\\
            \midrule
                \textit{Klause et al.} (2022) \cite{klause2022scale-norm} & ResNet-9 & GroupNorm & 2.89 & 65.6 \\
                \textit{Nasirigerdeh et al.} (2022) & ResNet-8 & KernelNorm & 2.0 & \textbf{68.08} \\
                \textit{Ours} & KNResNet-13 & KernelNorm &  2.0 & \textbf{68.05}$\pm$0.07 \\

            \bottomrule
            \end{tabular}
        }
    \end{minipage}
    \par
    \vskip 0.2in
    \centering
    \begin{minipage}{.8\textwidth}
        \subcaption{CIFAR-10 with augmentation multiplicity (K)}
        \vskip -0.05in
        \resizebox{\columnwidth}{!}{\begin{tabular}{llllll}
            \toprule
             Study & Model & Normalization & K & $\varepsilon$ & Test accuracy \\
            \midrule
                \textit{De  et al.} (2022) \cite{de2022dp-deepmind} & Wide ResNet-16-4 & GroupNorm & 16 & 8.0 & 79.5 \\
                \textit{De  et al.} (2022) \cite{de2022dp-deepmind} & Wide ResNet-40-4 & GroupNorm & 32 & 8.0 & \textbf{81.4} \\
                \textit{Ours} & KNResNet-13 & KernelNorm & 3 & 8.0 & 80.8 $\pm$0.22 \\
                
            \midrule
                \textit{De  et al.} (2022) \cite{de2022dp-deepmind} & Wide ResNet-16-4 & GroupNorm & 16 & 6.0 & 77.0 \\
                \textit{De  et al.} (2022) \cite{de2022dp-deepmind} & Wide ResNet-40-4 & GroupNorm & 32 & 6.0 & 78.8 \\
                \textit{Ours} & KNResNet-13 & KernelNorm & 3 & 6.0 & \textbf{79.09}$\pm$0.07 \\
                
            \midrule
                \textit{De  et al.} (2022) \cite{de2022dp-deepmind} & Wide ResNet-16-4 & GroupNorm & 16 & 4.0 & 71.9 \\
                \textit{De  et al.} (2022) \cite{de2022dp-deepmind} & Wide ResNet-40-4 & GroupNorm & 32 & 4.0 & 73.5 \\
                \textit{Ours} & KNResNet-13 & KernelNorm & 3 & 4.0 & \textbf{76.19}$\pm$0.04 \\
                
            \midrule
                \textit{De  et al.} (2022) \cite{de2022dp-deepmind} & Wide ResNet-16-4 & GroupNorm & 16 & 2.0 & 64.9 \\
                \textit{De  et al.} (2022) \cite{de2022dp-deepmind} & Wide ResNet-40-4 & GroupNorm & 32 & 2.0 & 65.9 \\
                \textit{Ours} & KNResNet-13 & KernelNorm & 3 & 2.0 & \textbf{70.57}$\pm$0.24 \\
            \bottomrule
            \end{tabular}
        }
    \end{minipage}
    \par
    \vskip 0.2in
    \centering
    \begin{minipage}{.8\textwidth}
        \subcaption{Imagenette}
        \vskip -0.05in
        \resizebox{\columnwidth}{!}{\begin{tabular}{llllll}
            \toprule
             Study & Model & Normalization & $\varepsilon$ & Test accuracy \\
            \midrule
                 \textit{Klause et al.} (2022) \cite{klause2022scale-norm} & ResNet-9 & GroupNorm & 7.42 & 64.8 \\
                 \textit{Klause et al.} (2022) \cite{klause2022scale-norm} & ResNet-9 & GroupNorm & 9.88 & 67.1 \\
                 \textit{Remerscheid et al.} (2022) \cite{remerscheid2022smoothnet} & DenseNet-14 & GroupNorm & 7.0 & 69.7 \\
                 \textit{Ours} & KNResNet-13 & KernelNorm & 7.0 & \textbf{72.24}$\pm$0.48 \\
            \bottomrule
            \end{tabular}
        }
    \end{minipage}
\vskip -0.1in
\end{table*}

In the following, we describe the data preprocessing and differentially private training procedure for the CIFAR-10 and Imagenette datasets. Then, we provide the accuracy values achieved by the KNResNet-13 model and compare them with those from the recent studies.

\textbf{CIFAR-10.} The only data preprocessing step is to divide the feature values by $255$. KNResNet-13 is trained for $T$ = $50$, $70$, $70$, and $80$ epochs with batch sizes of $B$=$4096$, $4096$, $3072$, and $3072$ for $\varepsilon$=$2.0$, $4.0$, $6.0$, and $8.0$, respectively. The learning rate is $2.0$, clipping value is $1.5$, and $\delta$ is $10^{-5}$. The learning rate is divided by $2$ at epochs (T - 30) and (T - 10). The optimizer is SGD with momentum of zero.

\textbf{CIFAR-10 with augmentation multiplicity.} The augmentation multiplicity is a recently proposed technique by \textit{De et al.} \cite{de2022dp-deepmind}, which computes the gradients for a given sample by taking average over the gradients computed for different augmentations of the same sample. For the CIFAR-10 dataset, this technique applies the sequence of random horizontal flipping and random cropping of size $32 \times 32$ and padding $4 \times 4$ to obtain an augmented version of a given sample. Here, we employ a slightly different way of augmentation multiplicity because the original technique provides negligible accuracy gain for our model. We first compute the gradients for the original sample, horizontally flipped (i.e. with probability of $1.0$), and randomly cropped version of the sample, and then take the average over them to calculate the per-sample gradients. For $\varepsilon$=$2.0$, $4.0$, $6.0$, and $8.0$, KNResNet-13 is trained for $80$, $80$, $100$, and $100$ epochs, respectively. The other training details are the same as CIFAR-10 with no augmentation multiplicity (previous paragraph).

\textbf{Imagenette. } We adopt the $320$-pixel version of the dataset and resize the images to $224 \times 224$. We train  KNResNet-13 with $\eta$=$1.5$, $C$=$1.5$, $\varepsilon$=$7.0$, $\delta$=$10^{-5}$, and zero-momentum SGD for $100$ epochs, where $\eta$ is divided by $2$ at epochs $70$ and $90$.

\textbf{Results.} Table \ref{tab:accuracy-comparision} lists the test accuracy values from KNResNet-13 and the recent studies on CIFAR-10, CIFAR-10 with augmentation multiplicity, and Imagenette. KNResNet-13 delivers significantly higher accuracy than the models based on GroupNorm or NoNorm for all considered $\varepsilon$ values on CIFAR-10 without augmentation multiplicity. Compared to kernel normalized ResNet-8 \cite{nasirigerdeh2022kernel-norm}, KNResNet-13 provides up to $2\%$ accuracy gain depending on the $\varepsilon$ value.

On CIFAR-10 with augmentation multiplicity, KNResNet-13 outperforms both wide ResNet-16-4 and ResNet-40-4  \cite{zagoruyko2016wide-resnet} with much lower augmentation multiplicity ($3$ vs. $16$ vs. $32$) for $\varepsilon$ values of $2.0$, $4.0$, and $6.0$. On Imagenette, KNResNet-13 achieves around $3\%$ and $7\%$ higher accuracy than GroupNorm based DenseNet-14 \cite{remerscheid2022smoothnet} and ResNet-9 \cite{klause2022scale-norm}, respectively. 

Given the results from Table \ref{tab:accuracy-comparision}, we provide new SOTA accuracy values on the CIFAR-10 and Imagenette datasets, when trained from scratch:
\begin{itemize}
    \item On CIFAR-10 \textit{without} augmentation multiplicity, the accuracy values of $74.51\%$, $77.09\%$, and $78.51\%$ for $\varepsilon$=$4.0$, $6.0$, and $8.0$, respectively.  
    \item On CIFAR-10 \textit{with} augmentation multiplicity, the accuracy values of $70.57\%$, $76.19\%$, and $79.09\%$ for $\varepsilon$=$2.0$, $4.0$, and $6.0$, respectively. 
    \item On Imagenette, the accuracy value of $72.24\%$ for $\varepsilon$=$7.0$.
\end{itemize}

\section{Discussion}
\label{sec:discussion}
Our experimental evaluation shows KernelNorm delivers higher performance than LayerNorm and GroupNorm in FL, DP, and DP-FL. This can be because KernelNorm is a local normalization method, taking into account the spatial correlation of the elements in the height and width dimensions during normalization. This leads to faster convergence rate compared to global batch-independent layers including LayerNorm and GroupNorm, likely due to the smoother optimization landscape \cite{klause2022scale-norm}. It implies KernelNorm requires less amount of total injected noise to achieve a target accuracy value for a given privacy budget in DP, and a fewer number of communication rounds, and thus, higher communication efficiency in FL.

Moreover, LayerNorm and GroupNorm have scale and shift as learnable parameters. In FL these parameters are aggregated, while they are perturbed with noise in DP. The performance of the layer and group normalized models can negatively be impacted in both cases. KernelNorm, however, is free of these learnable parameters, which can be another factor in superior performance of KernelNorm compared to LayerNorm and GroupNorm. 

Finally, the feature values are not required to be normalized with the per-channel mean and standard deviation of the dataset in KernelNorm based models due to self-normalizing nature of KNConv, which normalizes the input before computing convolution. This is beneficial, especially in federated environments, because it is not required for clients to share the mean and standard deviation of their local datasets with server to compute the corresponding global values. 

Given the aforementioned properties and its superior performance, KernelNorm has a great potential  to become the standard normalization layer for federated learning, differential privacy, and differentially private federated learning.

\section{Related Work}
\label{sec:related_work}
There are few studies that compare the performance of various normalization layers in federated settings. \textit{Hsieh et al.} \cite{hsieh2020fl-non-iid} experimentally show GroupNorm delivers higher accuracy than BatchNorm in supervised FL. \textit{Zhang et al.} \cite{zhang2020groupnorm-vs-batchnorm} demonstrate this also holds for semi-supervised FL. However, these studies have not compared GroupNorm with NoNorm as the baseline. Our experiments illustrate GroupNorm does not necessarily provide accuracy gain compared to NoNorm for shallow models in supervised federated settings. 

Several studies investigate the performance of different batch-independent normalization layers for differentially private learning. \textit{Klause et al.} \cite{klause2022scale-norm} and \textit{Remerscheid et al.} \cite{remerscheid2022smoothnet} show GroupNorm outperforms LayerNorm in terms of accuracy in DP settings. \textit{Nasirigerdeh et al.} \cite{nasirigerdeh2022kernel-norm} illustrate KernelNorm delivers considerable accuracy gain compared to both LayerNorm and GroupNorm in DP. These prior works, however, do not consider NoNorm as the baseline for comparison. Our evaluation indicates NoNorm slightly outperforms both LayerNorm and GroupNorm for the shallow ResNet-8 model on CIFAR-10, whereas KernelNorm still provides significant accuracy improvement compared to NoNorm for the aforementioned setting. The experimental evaluation of \textit{Nasirigerdeh et al.} \cite{nasirigerdeh2022kernel-norm}, moreover, is limited to a single case study. We conduct more extensive experiments with deeper models on both low-resolution and medium-resolution datasets to draw the performance comparisons among NoNorm, LayerNorm, GroupNorm, and KernelNorm.

Some studies propose novel architectures or data augmentation techniques to enhance the accuracy of differentially private models. \textit{Klause et al.} \cite{klause2022scale-norm} present a 9-layer ResNet architecture in which an additional normalization is performed after the addition operation of the residual block, and show their architecture improves the accuracy compared to the original ResNet architecture. \textit{Remerscheid et al.} \cite{remerscheid2022smoothnet} introduce a novel DenseNet-based architecture called SmoothNet, which employs $3 \times 3$ convolutional layers with a high number of filters in the DenseNet blocks, and demonstrate it outperforms the previous ones in terms of accuracy. Both architectures employ GroupNorm as their normalization layer. We propose the KNResNet-13 architecture based on KernelNorm, and show it delivers considerably higher accuracy than the aforementioned architectures on CIFAR-10 and Imagenette.

\textit{De et al.} \cite{de2022dp-deepmind} present the augmentation multiplicity technique, which computes the per-sample gradients by taking average over the gradients from different augmentations of the sample. We adopt this technique to train the proposed KNResNet-13 architecture on CIFAR-10. The accuracy from KNResNet-13 is higher than the wide ResNet-16-4 and ResNet-40-4 used in \cite{de2022dp-deepmind} for $\varepsilon$ values of $2.0$, $4.0$, and $6.0$.

\section{Conclusion and Future Work}
\label{sec:conclusion}
We address the normalization challenge in the context of federated and differentially private learning. Through extensive experiments, we demonstrate: (1) in FL and DP, using no normalization layer in the architecture of shallow networks such as VGG-6 and ResNet-8 delivers slightly higher accuracy than LayerNorm and GroupNorm, (2) on deeper models such as DenseNet-20$\times$16 and PreactResNet-18 in FL and DP as well as the shallow models in DP-FL, however, LayerNorm and GroupNorm considerably outperform NoNorm, and (3) the recently proposed KernelNorm method achieves significantly higher accuracy and convergence rate compared to NoNorm, LayerNorm, and GroupNorm in FL, DP, and DP-FL.

Given the superior performance of KernelNorm, we propose a kernel normalized ResNet architecture called KNResNet-13 for differentially private learning. Using the proposed architecture, we provide new SOTA accuracy values on CIFAR-10 with and without augmentation multiplicity as well as Imagenette for different $\varepsilon$ values, when trained from scratch. 

We employ a low augmentation multiplicity value  (i.e. $3$) in our study due to the remarkable computational overhead of the technique. KNResNet-13 might deliver even higher accuracy with larger augmentation multiplicity values (e.g. $16$ or $32$), which can be an investigated in future studies. Additionally, the performance evaluation of kernel normalized architectures on the large Imagenet-32$\times$32 dataset \cite{deng2009-imagenet} is an interesting direction for future works.

\bibliography{main}
\bibliographystyle{unsrt}

\clearpage
\onecolumn
\appendix
\begin{table}[!ht]
\vskip 0.35in
\caption{\textbf{Federated learning}: Learning rate values giving the highest accuracy for each normalization layer; B: batch size.}
\vskip -0.05in
\label{tab:lr-bs-fl}
    \begin{minipage}{.46\textwidth}
        \centering
        \subcaption{CIFAR-10-VGG-6 \textbf{(cross-silo FL)}}
        \vskip -0.1in
        \resizebox{\columnwidth}{!}{\begin{tabular}{lcccc}
            \toprule
             B & NoNorm& LayerNorm & GroupNorm & KernelNorm \\
            \midrule
            16  & 0.025 & 0.025 & 0.01 & 0.025 \\
            64  & 0.025 & 0.025 & 0.05 & 0.025 \\
            \bottomrule
            \end{tabular}
        }
    \end{minipage}
    \hspace{0.02\textwidth}
    \begin{minipage}{.46\textwidth}
        \centering
        \subcaption{CIFAR-10-VGG-6 \textbf{(cross-device FL)}}
        \vskip -0.1in
        \resizebox{\columnwidth}{!}{\begin{tabular}{lcccc}
            \toprule
             B & NoNorm & LayerNorm & GroupNorm & KernelNorm \\
            \midrule
                16  & 0.025 & 0.025 & 0.05 & 0.025 \\
                64  & 0.05 & 0.025 & 0.05 & 0.05 \\
            \bottomrule
            \end{tabular}
        }
    \end{minipage}
    \par
    \vskip 0.1in
    \centering
    \begin{minipage}{.46\textwidth}
        \centering
        \subcaption{CIFAR-100-PreactResNet-18 \textbf{(cross-silo FL)}}
        \vskip -0.1in
        \resizebox{\columnwidth}{!}{\begin{tabular}{lcccc}
            \toprule
             B & NoNorm & LayerNorm & GroupNorm & KernelNorm \\
            \midrule
            16  & 0.01 & 0.01 & 0.005 & 0.025 \\
            64  & 0.01 & 0.01 & 0.01 & 0.05 \\
            \bottomrule
            \end{tabular}
        }
    \end{minipage}
    \hspace{0.02\textwidth}
    \begin{minipage}{.46\textwidth}
        \centering
        \subcaption{CIFAR-100-PreactResNet-18 \textbf{(cross-device FL)}}
        \vskip -0.1in
        \resizebox{\columnwidth}{!}{\begin{tabular}{lcccc}
            \toprule
             B & NoNorm& LayerNorm & GroupNorm & KernelNorm \\
            \midrule
                16  & 0.01 & 0.01 & 0.005 & 0.025 \\
                64  & 0.05 & 0.01 & 0.01 & 0.1 \\
            \bottomrule
            \end{tabular}
        }
    \end{minipage}
\vskip -0.1in
\end{table}

\begin{table*}[!ht]
\caption{\textbf{Differential privacy}:  Learning rate values giving the highest accuracy for each normalization layer; B: batch size.}
\vskip -0.05in
\label{tab:lr-bs-dp}
    \begin{minipage}{.46\textwidth}
        \centering
        \subcaption{CIFAR-10-ResNet-8 (DP)}
        \vskip -0.1in
        \resizebox{\columnwidth}{!}{\begin{tabular}{lllll}
            \toprule
             B & NoNorm& LayerNorm & GroupNorm & KernelNorm \\
            \midrule
                512 & 1.0 & 1.0 & 1.0 & 1.0 \\
                1024 & 2.0 & 2.0 & 1.5 & 1.5 \\
                2048 & 2.0 & 2.0 & 2.0 & 2.0 \\
                3072 & 2.0 & 2.0 & 2.0 & 2.0 \\
            \bottomrule
            \end{tabular}
        }
    \end{minipage}
    \hspace{0.02\textwidth}
    \begin{minipage}{.46\textwidth}
        \centering
        \subcaption{CIFAR-10-DenseNet-20$\times$16 (DP)}
        \vskip -0.1in
        \resizebox{\columnwidth}{!}{\begin{tabular}{lcccc}
            \toprule
             B & NoNorm & LayerNorm & GroupNorm & KernelNorm \\
            \midrule
            256 & 1.0 & 1.5 & 2.0 & 1.5 \\
            512 & 1.0 & 2.0 & 2.0 & 1.5 \\
            1024 & 1.5 & 2.0 & 1.5 & 1.5 \\
            2048 & 2.0 & 2.0 & 2.0 & 1.5 \\
            \bottomrule
            \end{tabular}
        }
    \end{minipage}
    \par
    \vskip 0.1in
    \centering
    \begin{minipage}{.46\textwidth}
        \centering
        \subcaption{Imagenette-PreactResNet-18 (DP)}
        \vskip -0.1in
        \resizebox{\columnwidth}{!}{\begin{tabular}{lcccc}
            \toprule
             B & NoNorm& LayerNorm & GroupNorm & KernelNorm \\
            \midrule
                512 & 1.0 & 1.0 & 1.0 & 1.5 \\
                1024 & 1.0 & 1.0 & 1.0 & 2.0 \\
                2048 & 1.5 & 1.0 & 1.0 & 2.0 \\
            \bottomrule
            \end{tabular}
        }
    \end{minipage}
\vskip -0.1in
\end{table*}

\begin{table*}[!ht]
\vskip -0.05in
\caption{\textbf{Differential privacy}:  Clipping values giving the highest accuracy for each normalization layer; B: batch size.}
\label{tab:clipping-bs-dp}
    \begin{minipage}{.46\textwidth}
        \centering
        \subcaption{CIFAR-10-ResNet-8 (DP)}
        \vskip -0.1in
        \resizebox{\columnwidth}{!}{\begin{tabular}{lllll}
            \toprule
             B & NoNorm& LayerNorm & GroupNorm & KernelNorm \\
            \midrule
                512 & 1.0 & 1.0 & 1.0 & 1.0 \\
                1024 & 1.0 & 1.5 & 2.0 & 1.5 \\
                2048 & 2.0 & 2.0 & 2.0 & 2.0 \\
                3072 & 2.0 & 2.0 & 2.0 & 2.0 \\
            \bottomrule
            \end{tabular}
        }
    \end{minipage}
    \hspace{0.02\textwidth}
    \begin{minipage}{.46\textwidth}
        \centering
        \subcaption{CIFAR-10-DenseNet-20$\times$16 (DP)}
        \vskip -0.1in
        \resizebox{\columnwidth}{!}{\begin{tabular}{lcccc}
            \toprule
             B & NoNorm & LayerNorm & GroupNorm & KernelNorm \\
            \midrule
            256 & 1.0 & 1.5 & 2.0 & 1.5 \\
            512 & 1.0 & 1.5 & 1.5 & 1.5 \\
            1024 & 2.0 & 2.0 & 2.0 & 1.5 \\
            2048 & 2.0 & 1.5 & 2.0 & 1.0 \\
            \bottomrule
            \end{tabular}
        }
    \end{minipage}
    \par
    \vskip 0.15in
    \centering
    \begin{minipage}{.46\textwidth}
        \centering
        \subcaption{Imagenette-PreactResNet-18 (DP)}
        \vskip -0.1in
        \resizebox{\columnwidth}{!}{\begin{tabular}{lcccc}
            \toprule
             B & NoNorm& LayerNorm & GroupNorm & KernelNorm \\
            \midrule
                512 & 1.0 & 1.0 & 1.0 & 1.5 \\
                1024 & 1.0 & 1.5 & 1.0 & 1.0 \\
                2048 & 1.0 & 1.0 & 1.0 & 1.0 \\
            \bottomrule
            \end{tabular}
        }
    \end{minipage}
\vskip -0.1in
\end{table*}

\begin{table*}[!ht]
\vskip -0.05in
\caption{\textbf{Differentially private federated learning}:  Learning rates giving the highest accuracy for each norm layer.}
\label{tab:lr-bs-dp-fl}
    \begin{minipage}{.46\textwidth}
        \centering
        \subcaption{CIFAR-10-VGG-6 (DP-FL)}
        \vskip -0.1in
        \resizebox{\columnwidth}{!}{\begin{tabular}{lllll}
            \toprule
             B & NoNorm& LayerNorm & GroupNorm & KernelNorm \\
            \midrule
                256 & 0.01 & 0.01 & 0.01 & 0.01 \\
                512 & 0.025 & 0.01 & 0.01 & 0.025 \\
                1024 & 0.025 & 0.01 & 0.025 & 0.025 \\
            \bottomrule
            \end{tabular}
        }
    \end{minipage}
    \hspace{0.02\textwidth}
    \begin{minipage}{.46\textwidth}
        \centering
        \subcaption{CIFAR-10-ResNet-8 (DP-FL)}
        \vskip -0.1in
        \resizebox{\columnwidth}{!}{\begin{tabular}{lcccc}
            \toprule
             B & NoNorm & LayerNorm & GroupNorm & KernelNorm \\
            \midrule
                256 & 0.01  & 0.01 & 0.01 & 0.01 \\
                512 & 0.025 & 0.01 & 0.01 & 0.01 \\
                1024 & 0.025 & 0.01 & 0.01 & 0.05 \\
            \bottomrule
            \end{tabular}
        }
    \end{minipage}
\vskip -0.1in
\end{table*}

\begin{table*}[!ht]
\vskip -0.05in
\caption{\textbf{Differentially private federated learning}:  Clipping values giving the highest accuracy for each norm layer.}
\label{tab:clipping-bs-dp-fl}
    \begin{minipage}{.46\textwidth}
        \centering
        \subcaption{CIFAR-10-VGG-6 (DP-FL)}
        \vskip -0.1in
        \resizebox{\columnwidth}{!}{\begin{tabular}{lllll}
            \toprule
             B & NoNorm& LayerNorm & GroupNorm & KernelNorm \\
            \midrule
                256 & 1.0 & 1.0 & 1.5 & 1.0 \\
                512 & 1.5 & 1.0 & 1.0 & 1.0 \\
                1024 & 2.0 & 1.5 & 2.0 & 2.0 \\
            \bottomrule
            \end{tabular}
        }
    \end{minipage}
    \hspace{0.02\textwidth}
    \begin{minipage}{.46\textwidth}
        \centering
        \subcaption{CIFAR-10-ResNet-8 (DP-FL)}
        \vskip -0.1in
        \resizebox{\columnwidth}{!}{\begin{tabular}{lcccc}
            \toprule
             B & NoNorm & LayerNorm & GroupNorm & KernelNorm \\
            \midrule
                256 & 1.0  & 1.5 & 1.0 & 1.0 \\
                512 & 1.0 & 1.0 & 1.0 & 1.0 \\
                1024 & 1.0 & 1.0 & 2.0 & 2.0 \\
            \bottomrule
            \end{tabular}
        }
    \end{minipage}
\vskip -0.1in
\end{table*}

\end{document}